\pdfoutput=1

\documentclass[11pt]{article}

\usepackage[]{acl}

\usepackage{times}
\usepackage{latexsym}
\usepackage{textcomp}
\usepackage{float}
\usepackage{multirow}
\usepackage{booktabs}
\usepackage{graphicx}
\usepackage[T1]{fontenc}
\usepackage{courier}
\usepackage{mathtools}

\usepackage[utf8]{inputenc}
\usepackage{graphicx}
\usepackage{subfig}
\usepackage[belowskip=-5pt,aboveskip=5pt]{caption}

\usepackage{microtype}
\usepackage{inconsolata}
\usepackage[figuresright]{rotating}

\usepackage[T1]{fontenc}
\usepackage{tgcursor}
\usepackage{xcolor}
\usepackage{xspace}
\usepackage[utf8]{inputenc}
\usepackage{authblk}
\usepackage{natbib}
\usepackage{lipsum}
\usepackage{multirow}
\usepackage{enumitem}

\title{Text-Augmented Open Knowledge Graph Completion via \\ Pre-Trained Language Models}
\author[ ]{\textbf{Pengcheng Jiang}}
\author[ ]{\textbf{Shivam Agarwal}}
\author[ ]{\textbf{Bowen Jin}}
\author[ ]{\textbf{Xuan Wang}$^{\dagger}$}
\author[ ]{\\\textbf{Jimeng Sun}}
\author[ ]{\textbf{Jiawei Han}}
\affil[ ]{Department of Computer Science, University of Illinois at Urbana-Champaign}
\affil[ ]{$^{\dagger}$Department of Computer Science, Virginia Tech}
\affil[ ]{\texttt{\{pj20, shivama2, bowenj4, jimeng, hanj\}@illinois.edu} \quad \texttt{xuanw@vt.edu}}

\newcommand\mask{\texttt{[MASK]}}

\date{}

\begin{document}
\maketitle
\begin{abstract}
The mission of open knowledge graph (KG) completion is to draw new findings from known facts. Existing works that augment KG completion require either (1) factual triples to enlarge the graph reasoning space or (2) manually designed prompts to extract knowledge from a pre-trained language model (PLM), exhibiting limited performance and requiring expensive efforts from experts. To this end, we propose \textsc{TagReal} that automatically generates quality  query prompts and retrieves support information from large text corpora to probe knowledge from PLM for KG completion.
The results show that \textsc{TagReal} achieves state-of-the-art performance 
on two benchmark datasets.
We find that \textsc{TagReal} has superb performance even with limited training data, outperforming existing embedding-based, graph-based, and PLM-based methods.

\end{abstract}

\section{Introduction}

\begin{figure}[!h]

\centering
\includegraphics[width=\linewidth]{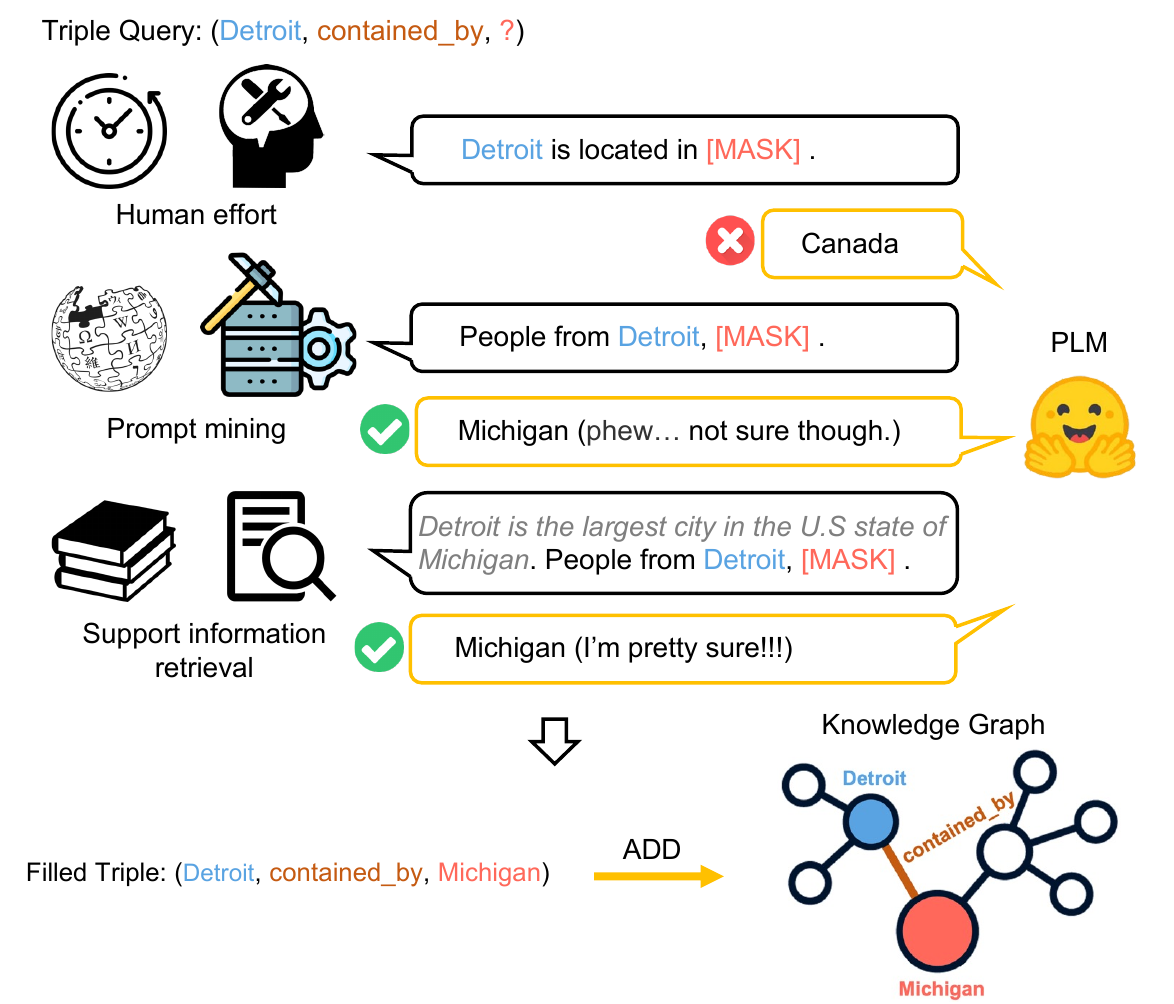}

\caption{The quality of hand-crafted prompts can be limited, while prompt mining is a scalable alternative. Support information also helps PLM understand the purpose of prompts. In this example, Canada and Michigan are potentially valid options but given prompt mining and support information retrieval, the model becomes confident about Michigan as the answer here. }
\label{fig:intro}
\vskip -2ex
\end{figure}

A knowledge graph (KG) is a heterogeneous graph that encodes factual information in the form of entity-relation-entity triplets, where a \textit{relation} connects a \textit{head} entity and a \textit{tail} entity (e.g., ``\textit{Miami-located\_in-USA}'') \cite{Wang2017KnowledgeGE,10.1145/3447772}.
KG \cite{Dai2020ASO} plays a central role in many NLP applications, including question answering \cite{hao-etal-2017-end,yasunaga-etal-2021-qa}, recommender systems \cite{Zhou2020InteractiveRS}, and drug discovery \cite{10.1093/bioinformatics/bty294}. 
However, existing works \cite{wang2018towards, hamilton2018embedding} show that most large-scale KGs are incomplete and cannot fully cover the massive real-world knowledge. 
This challenge motivates KG completion, which aims to find one or more object entities given a subject entity and a relation~\cite{Lin2015LearningEA}. For example, in Figure \ref{fig:intro}, our goal is to predict the object entity with ``\textit{Detroit}'' as the subject entity and ``\textit{contained\_by}'' as the relation.

However, existing KG completion approaches \cite{pmlr-v48-trouillon16,das2018go} have several limitations \cite{fu-etal-2019-collaborative}.
First, their performance heavily depends on the density of the graph. 
They usually perform well on dense graphs with rich structural information but poorly on sparse graphs which are more common in real-world applications.
Second, previous methods (e.g., \citet{bordes2013translating}) assume a closed-world KG without considering vast open knowledge in the external resources. In fact, in many cases, a KG is usually associated with a rich text corpus \cite{bodenreider2004unified}, which contains a vast amount 
of factual data
not yet extracted. To overcome these challenges we investigate the task of open knowledge graph completion, 
where KG can be constructed using new facts from outside the KG.
Recent text-enriched solutions \cite{fu-etal-2019-collaborative} focus on using a pre-defined set of facts to enrich the knowledge graph. 
Nonetheless, the pre-defined set of facts is often noisy and constricted, that is, they do not provide sufficient information to efficiently update the KG.

Pre-trained language models (PLMs) \cite{devlin-etal-2019-bert,Liu2019RoBERTaAR} have shown to be powerful in capturing factual knowledge implicitly from learning on massive unlabeled texts \cite{petroni-etal-2019-language}. 
Since PLMs are superb in text encoding, 
they can be utilized to facilitate knowledge graph completion with external text information.
Recent knowledge graph completion methods \cite{shin-etal-2020-autoprompt,lv2022pre} focus on using manually crafted prompts (e.g., ``Detroit is located in [MASK]'' in Figure \ref{fig:intro}) to query the PLMs for graph completion (e.g., ``Michigan'').
However, manually creating prompts can be expensive with limited quality (e.g., PLM gives a wrong answer ``Canada'' to the query with a handcrafted prompt, as shown in Figure \ref{fig:intro}). 

Building on the above limitations of standard KG and the enormous power of PLMs \cite{devlin-etal-2019-bert,Liu2019RoBERTaAR}, we aim to use PLMs for open knowledge graph completion. 
We propose an end-to-end framework that jointly exploits the implicit knowledge in PLMs and textual information in the corpus to perform knowledge graph completion (as shown in Figure \ref{fig:intro}).
Unlike existing works (e.g., \cite{fu-etal-2019-collaborative, lv2022pre}), our method does not require a manually pre-defined set of facts and prompts, which is more general and easier to adapt to real-world applications.

 Our contributions can be summarized as: \\
\begin{itemize}
\vspace{-4ex}
    \item We study the open KG completion problem that can be assisted by facts captured from PLMs. To this end, we propose a new framework \textbf{\textsc{TagReal}} that denotes \textbf{t}ext \textbf{a}u\textbf{g}mented open KG completion with \textbf{real}-world knowledge in PLMs.
    \vspace{-2ex}
    \item We develop prompt generation and information retrieval methods, which enable \textsc{TagReal} to automatically create high-quality prompts for PLM knowledge probing and search support information, making it more practical especially when PLMs lack some domain knowledge.
    \vspace{-2ex}
    \item Through extensive quantitative and qualitative experiments on real-world knowledge graphs such as Freebase\footnote{\url{https://github.com/thunlp/OpenNRE}} we show the applicability and advantages of our framework\footnote{Our code is available at: \url{https://github.com/pat-jj/TagReal}}.
\end{itemize}

\section{Related Work}

\subsection{KG Completion Methods}

KG completion methods can be categorized into embedding-based and PLM-based methods.
\textbf{Embedding-based methods} represent entities and relations as embedding vectors and maintain their semantic relations in the vector space.
TransE \cite{bordes2013translating} vectorizes the head, the relation and the tail of triples into a Euclidean space.
DistMult \cite{yang2014embedding} converts all relation embeddings into diagonal matrices in bilinear models.
RotatE \cite{sun2018rotate} presents each relation embedding as a rotation in complex vector space from the head entity to the tail entity. 

\indent In recent years, researchers have realized that PLMs can serve as knowledge bases \cite{petroni2019language, zhang2020empower,alkhamissi2022review}. 
\textbf{PLM-based methods} for KG completion \cite{yao2019kg, kim2020multi, chang2021incorporating, lv2022pre} start to gain attention. 
As a pioneer, KG-BERT \cite{yao2019kg} fine-tunes PLM with concatenated head, relation, and tail in each triple, outperforming the conventional embedding-based methods in link prediction tasks. 
\citeauthor{lv2022pre}\citeyearpar{lv2022pre} present PKGC, which uses manually designed triple prompts and carefully selected support prompts as inputs to the PLM.
Their result shows that PLMs could be used to substantially improve the KG completion performance, especially in the \textit{open-world} \cite{shi2018open} setting. Compared to PKGC, our framework \textsc{TagReal} automatically generates prompts of higher quality without any domain expert knowledge. Furthermore, instead of pre-supposing the existence of support information, we search relevant textual information from the corpus with an information retrieval method to support the PLM knowledge probing.

\subsection{Knowledge Probing using Prompts}
LAMA \cite{petroni2019language} is the first framework for knowledge probing from PLMs. The prompts are manually created with a subject placeholder and an unfilled space for the object. For example, a triple query (\textit{Miami}, \textit{location}, ?) may have a prompt “Miami is located in \mask” where “\texttt{<subject>} is located in \mask” is the template for ``location'' relation. The training goal is to correctly fill \mask with PLM’s prediction. Another work, BertNet \cite{hao2022bertnet}, proposes an approach applying GPT-3 \cite{brown2020language} to automatically generate a weighted prompt ensemble with input entity pairs and a manual seed prompt. It then uses PLM again to search and select top-ranked entity pairs 
with the ensemble
for KG completion. 

\subsection{Prompt Mining Methods}
When there are several relations to interpret, manual prompt design is costly due to the requirement of domain expert knowledge. In addition, the prompt quality could not be ensured. Hence, {\it quality prompt mining} catches the interest of researchers. \citeauthor{jiang2020can} \citeyear{jiang2020can} propose an approach MINE which searches middle words or dependency paths between the given inputs and outputs in a large text corpus (e.g., Wikipedia). They also propose a reasonable approach to optimize the ensemble of the mined prompts by weighting prompt individuals regarding their performance on the PLM. \\
\indent Before the emergence and widespread use of PLMs, textual pattern mining performed a similar function to find reliable patterns for information extraction. For instance, MetaPAD \cite{jiang2017metapad} generates quality meta patterns by context-aware segmentation with the pattern quality function, and TruePIE \cite{li2018truepie} 
proposes the concept of pattern embedding and a self-training framework, that discovers positive patterns automatically.

\begin{figure}[t]

\centering
\includegraphics[width=0.49\textwidth]{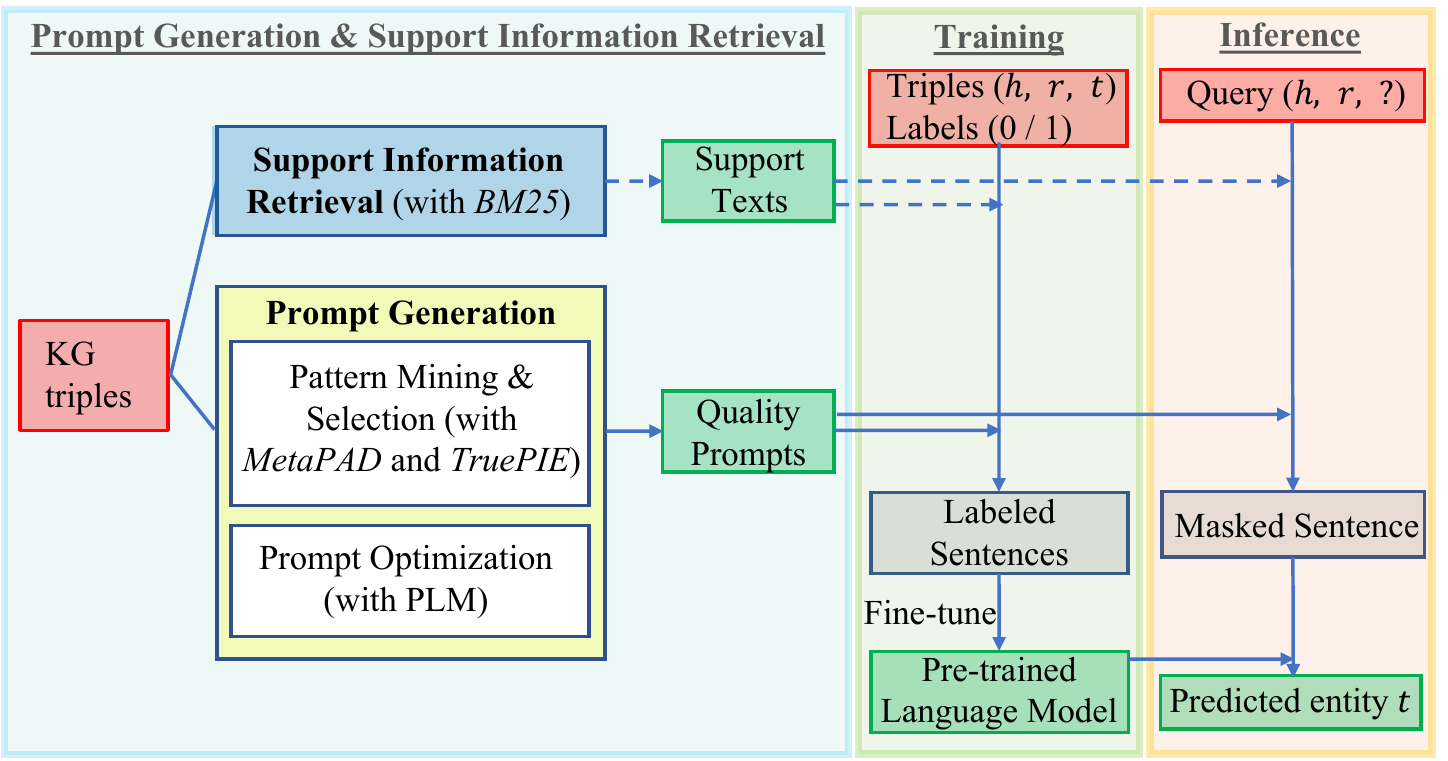}
\caption{ \textbf{\textsc{TagReal} Framework}. The input and output of each phase are highlighted by red and green, respectively. The dotted arrow indicates the optional process.}
\label{fig:framework}
\vskip -2ex
\end{figure}


\section{Methodology}
\begin{figure*}[!h]
\definecolor{green}{HTML}{2ECC71}
\centering
  \includegraphics[width=15.68cm,height=7cm]{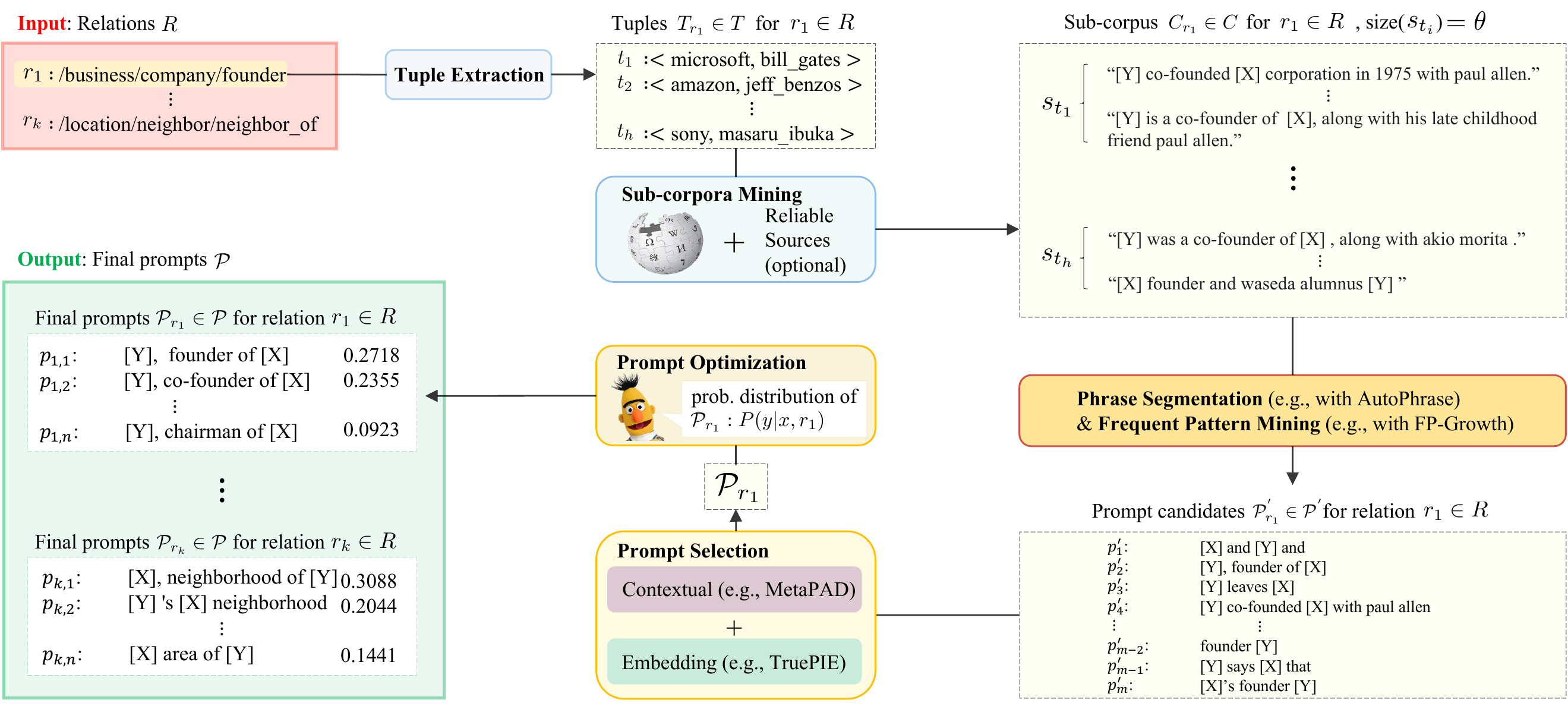}
  \caption{\textbf{Prompt generation process}. The solid lines connect the intermediate processes, and the arrows point to the intermediate/final results. Input and output are highlighted in \textbf{{\color{red} red}} and \textbf{{\color{green} green}} respectively. \texttt{[X]} and \texttt{[Y]} denote head and tail entities respectively.}
  \label{fig:pm}
  \vskip -1ex
\end{figure*}

We propose \textsc{TagReal}, a PLM-based framework to handle KG completion tasks. In contrast to the previous work, our framework does not rely on handcrafted prompts or pre-defined relevant facts. As shown in Figure \ref{fig:framework}, 
we automatically create appropriate prompts and search relevant support information,
which are further utilized as templates to explore implicit knowledge from PLMs.
\subsection{Problem Formulation}
\textbf{Knowledge graph completion} is to add new triples (facts) to the existing triple set of a KG. There are two tasks to achieve this goal. The first is \textbf{triple classification}, which is a binary classification task to predict whether a triple \((h,r,t)\) belongs to the KG, where \(h,r,t\) denote head entity, relation and tail entity respectively. The second task is \textbf{link prediction}, which targets on predicting either the tail entity \(t\) with a query \((h,r,?)\) or the head entity \(h\) with a query \((?,r,t)\).

\subsection{Prompt Generation}
\label{sec:3.1}
Previous studies (e.g., \citet{jiang2020can}) demonstrate that the accuracy of relational knowledge extracted from PLMs heavily relies on the quality of prompts used for querying. To this end, we develop a comprehensive approach for automatic quality prompt generation given triples in KG as the only input, as shown in Figure \ref{fig:pm}. We use textual pattern mining methods to mine quality patterns from large corpora as the prompts used for PLM knowledge probing. 
As far as we know, we are pioneers in using \textbf{textual pattern mining} methods for \textbf{LM prompt mining}.
We believe in the applicability of this approach for the following reasons.
\begin{itemize}[leftmargin=*]
\setlength\itemsep{0em}
    \item Similar data sources. We apply pattern mining on large corpora (e.g., Wikipedia) which are the data sources where most of 
    PLMs are pre-trained.
    \item Similar objectives. Textual pattern mining is to mine patterns to extract new information from large corpora; prompt mining is to mine prompts to probe implicit knowledge from PLMs.
    \item Similar performance criteria. The reliability of a pattern or a prompt is indicated by how many accurate facts it can extract from corpora/PLMs.
\end{itemize}

\indent\textbf{Sub-corpora mining} is the first step that creates the data source for the pattern mining. Specifically, given a KG with a relation set \(R=(r_1,r_2,...,r_k)\), we first extract tuples \(T_{r_i}\) paired by head entities and tail entities for each relation \(r_i\in R\) from the KG. For example, for the relation \(r_1\): \texttt{/business/company/founder}, we extract all tuples like \texttt{<microsoft, bill\_gates>} in this relation from the KG. For each tuple \(t_j\), we then search sentences \(s_{t_j}\) containing both head and tail from a large corpus (e.g., Wikipedia) and other reliable sources, which is added to compose the sub-corpus \(C_{r_i}\). We limit the size of each set to \(\theta\) for each tuple to mine more generic patterns for future applications.

\textbf{Phrase segmentation and frequent pattern mining} are applied to mine patterns from sub-corpora as prompt candidates. We use AutoPhrase \cite{shang2018automated} to segment corpora to more natural and unambiguous semantic phrases,
and use FP-Growth algorithm \cite{han2000mining} to mine frequent appeared patterns to compose a candidate set \(\mathcal{P}_{r_i}^{'}=(p_{1}^{'},p_{2}^{'},...,p_{m}^{'})\). The size of the set is large, as there are plenty of messy textual patterns.

\textbf{Prompt selection}. To select quality patterns from the candidate set, we apply two textual mining approaches: MetaPAD \cite{jiang2017metapad} and TruePIE \cite{li2018truepie}. MetaPAD applies pattern quality function introducing several criteria of contextual features to estimate the reliability of a pattern. We explain why those features can also be adapted for LM prompt estimation: (1) \textit{Frequency and concordance}: Since a PLM learns more contextual relations between frequent patterns and entities during the pre-training stage, a pattern occurs more frequently in the background corpus can probe more facts from the PLM. Similarly, if a pattern composed of highly associated sub-patterns appears frequently, it should be considered as a good one as the PLM would be familiar with the contextual relations among the sub-patterns. (2) \textit{Informativeness}:  A pattern with low informativeness (e.g., \(p_{1}^{'}\) in Figure \ref{fig:pm}) has the weak ability of PLM knowledge probing, as the relation between the subject or object entities cannot be well interpreted by it. (3) \textit{Completeness}: The completeness of a pattern affects a lot to the PLM knowledge probing especially when any of the placeholders is missing (e.g., \(p_{m-2}^{'}\) in Figure \ref{fig:pm}) so that PLM cannot even give an answer. (4) \textit{Coverage}: A quality pattern should be able to probe accurate facts from PLM as many as possible. Therefore, patterns like \(p_{4}^{'}\) which only suit a few or only one case should have a low quality score. We then apply TruePIE on the prompts (patterns) selected by MetaPAD. TruePIE filters the prompts that have low cosine similarity with the positive samples (e.g., \(p_{3}^{'}\) and \(p_{m-1}^{'}\) are filtered), which matters to the creation of prompt ensemble since we want the prompts in the ensemble to be semantically close to each other so that some poor-quality prompts would not significantly impact the prediction result by PLM.
As a result, we create a more reliable prompt ensemble \(\mathcal{P}_{r_i} =\{p_{i,1}, p_{i,2},...,p_{i,n}\}\) 
based on the averaged probabilities given by the prompts:
\begin{equation}
\label{eq:1}
    P(y|x, r_{i}) = \frac{1}{n}\sum_{j=1}^{n}P_{LM}(y|x,p_{i,j}),
\end{equation}
\noindent where \(r_i\) is the \(i\)-th relation and \(p_{i,j}\) is the \(j\)-th prompt in \(\mathcal{P}_{r_i}\). Beyond prompt selection, a \textbf{prompt optimization} process is also employed. Pointed out by \citeauthor{jiang2020can} \citeyear{jiang2020can}, some prompts in the ensemble are more reliable and ought to be weighted more.
Thus, we change Equation \ref{eq:1} to:
\begin{equation}
\label{eq:2}
    P(y|x, r_{i}) = \sum_{j=1}^{n}w_{i,j}P_{LM}(y|x,p_{i,j}),
\end{equation}
\noindent where \(w_{i,j}\) is the weight of \(j\)-th prompt for \(i\)-th relation. In our setting, all weights \(\{w_{1,1},..,w_{k,n}\}\) are learned through PLM to optimize \(P(y|x,r_{i})\) for \(r_i \in R\) ahead of the training process.


\subsection{Support Information Retrieval}
\begin{figure}[t]
\centering
\includegraphics[width=0.8\linewidth]{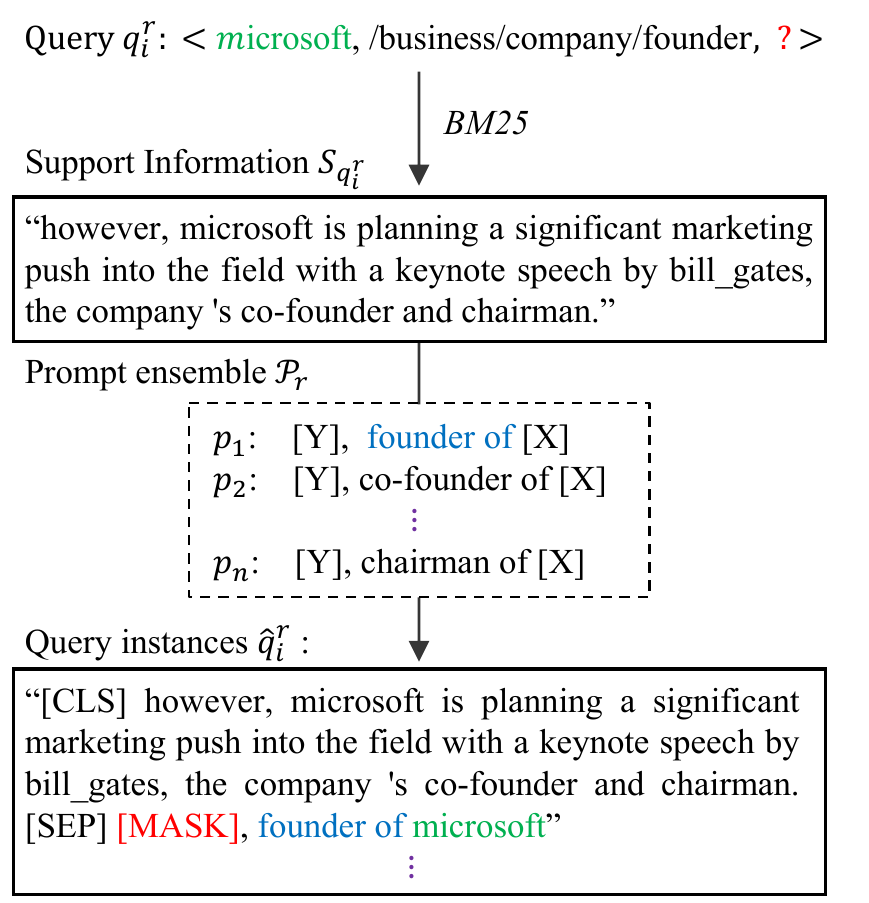}

\caption{Support information retrieval.}
\label{fig:sir}
\vskip -2ex
\end{figure}
In addition to the prompt mining, we also attach some query-wise and triple-wise support text information to the prompt to help the PLMs understand the knowledge we want to probe as well as to aid in training triple classification ability. As seen in Figure \ref{fig:sir}, for the \(i\)-th query \(q_i^r\) in relation \(r\), we use BM25 \cite{robertson1995okapi} to retrieve highly ranked support texts with score greater than \(\delta\) and length shorter than \(\phi\)  from the reliable corpus and randomly select one of them as the support information. To compose the input cloze \(\hat{q}_i^r\) to the PLM, we concatenate the support text to each prompt in the optimized ensemble we obtained through previous steps, with the subject filled and the object masked. \texttt{[CLS]} and \texttt{[SEP]} are the tokens for sequence classification and support information-prompt separation accordingly.

In the training stage, we search texts using triples rather than queries, and the \texttt{[MASK]} would be filled by the object entities. It is worth noting that support text is optional in \textsc{TagReal}, and we leave it blank if no matching data is found. 
\subsection{Training}
To train our model, we create negative triples in addition to the given positive triples following the idea introduced by PKGC \cite{lv2022pre}, to handle the triple classification task. We create negative triples by replacing the head and tail in each positive triple with the "incorrect" entity that achieves high probability by the KGE model.
We also create random negative samples by randomly replacing the heads and tails to enlarge the set of negative training/validation triples. The labeled training triples are assembled as \(\mathcal{T} = \mathcal{T}^{+} \cup (\mathcal{T}_{KGE}^{-} \cup \mathcal{T}_{RAND}^{-})\) where \(\mathcal{T}^{+}\) is the positive set,  \(\mathcal{T}_{KGE}^{-}\) and \( \mathcal{T}_{RAND}^{-}\) are two negative sets we created by embedding model-based and random approaches respectively. Then, we transform all training triples of each relation \(r\) into sentences with the prompt ensemble \(\mathcal{P}_{r}\) and the triple-wise support information retrieved by BM25 (if there is any). At the training stage, the \texttt{[MASK]} is replaced by the object entity in each positive/negative triple. The query instances \(\hat{q}_i^r\) are then used to fine-tune the PLM by updating its parameters. Cross-entropy loss \cite{lv2022pre} is applied for optimization:
\begin{equation}
\label{eq:3}
    \mathcal{L} = - \sum_{\tau \in \mathcal{T}}(y_{\tau}\log(c_{\tau}^{1}) + (1-y_{\tau})\frac{\log(c_{\tau}^{0})}{M}),
\end{equation}
\vskip -0.4em
\noindent where \(c_{\tau}^{0}, c_{\tau}^{1} \in [0,1]\) are the softmax classification scores of the token \texttt{[CLS]} for the triple \(\tau\), \(y_{\tau}\) is the ground truth label \((1/0)\) of the triple, and \(M = (|\mathcal{T}^{+}|/|\mathcal{T}^{-}|)\) is the ratio between the number of positive and negative triples. After the PLM is fine-tuned with positive/negative triples in training set, it should have a better performance on classifying the triples in the dataset compared to a raw PLM. This capability would enable it to perform KG completion as well.

\subsection{Inference}
\label{sec:3.4}
Given a query \((h,r,?)\), we apply the query-wise support information that is relevant to the head entity \(h\) and relation \(r\), as we presume that we are unaware of the tail entity (our prediction goal). Then, we make the corresponding query instances containing \texttt{[MASK]}, with both support information and prompt ensemble, as shown in Figure \ref{fig:sir}. To leverage the triple classification capability of the PLM on link prediction, we replace \texttt{[MASK]} in a query instance with each entity in the known entity set and rank their classification scores in descending order to create a 1-d vector as the prediction result for each query. This indicates that the lower-indexed entities in the vector are more likely to compose a positive triple with the input query. For prompt ensemble, we sum up the scores by entity index before ranking them. The detailed illustration is placed in Appendix \ref{ap:lp}.

\begin{table*}[!h]
\small
\centering
\setlength{\tabcolsep}{3.9pt}
\resizebox{\textwidth}{!}{

\begin{tabular}{clccccccccc}
\toprule
& Model      & \multicolumn{3}{c}{20\%}     & \multicolumn{3}{c}{50\%}     & \multicolumn{3}{c}{100\%}            \\       
&           
& Hits@5 & Hits@10 & MRR       & Hits@5 & Hits@10 & MRR       & Hits@5 & Hits@10 & MRR               \\    
\midrule
\midrule
\multirow{6}{*}{\textbf{KGE-based}} & 
TransE \cite{bordes2013translating}             & 29.13   & 32.67   & 15.80   & 41.54   & 45.74   & 25.82   & 42.53   & 46.77   & 29.86     \\
&
DisMult \cite{yang2014embedding}                & 3.44    & 4.31    & 2.64    & 15.98   & 18.85   & 13.14   & 37.94   & 41.62   & 30.56     \\
&
ComplEx \cite{trouillon2016complex}             & 4.32    & 5.48    & 3.16    & 15.00   & 17.73   & 12.21   & 35.42   & 38.85   & 28.59     \\
&
ConvE \cite{dettmers2018convolutional}          & 29.49   & 33.30   & 24.31   & 40.10   & 44.03   & 32.97   & 50.18   & 54.06   & 40.39     \\
&
TuckER \cite{balavzevic2019tucker}              & 29.50   & 32.48   & 24.44   & 41.73   & 45.58   & 33.84   & 51.09   & 54.80   & 40.47     \\
&
RotatE \cite{sun2018rotate}                     & 15.91   & 18.32   & 12.65   & 35.48   & 39.42   & 28.92   & \textbf{51.73}   & 55.27   & \textbf{42.64}     \\
\midrule
\multirow{3}{*}{\textbf{Text\&KGE-based}} & 
RC-Net \cite{xu2014rc}                          & 13.48   & 15.37   & 13.26   & 14.87   & 16.54   & 14.63   & 14.69   & 16.34   & 14.41     \\
& 
TransE+Line \cite{fu-etal-2019-collaborative}   & 12.17   & 15.16   & 4.88    & 21.70   & 25.75   & 8.81    & 26.76   & 31.65   & 10.97     \\
& 
JointNRE \cite{han2018neural}                   & 16.93   & 20.74   & 11.39   & 26.96   & 31.54   & 21.24   & 42.02   & 47.33   & 32.68     \\
\midrule
\multirow{2}{*}{\textbf{RL-based}} & 
MINERVA \cite{das2017go}                        & 11.64   & 14.16   & 8.93    & 25.16   & 31.54   & 22.24   & 43.80   & 44.70   & 34.62     \\
 & 
CPL \cite{fu-etal-2019-collaborative}           & 15.19   & 18.00   & 10.87   & 26.81   & 31.70   & 23.80   & 43.25   & 49.50   & 33.52     \\
\midrule
\multirow{2}{*}{\textbf{PLM-based}} & 
PKGC \cite{lv2022pre}                           & 35.77   & 43.82   & 28.62   & 41.93   & 46.70   & 31.81   & 41.98   & 52.56   & 32.11     \\ 
& 
TagReal (our method)                       & \textbf{45.59}   & \textbf{51.34}   & \textbf{35.41}   & \textbf{48.98}   & \textbf{55.64}   & \textbf{38.03}   & 50.85   & \textbf{60.64}   & 38.86     \\ 
\bottomrule
\end{tabular}
}
\caption{\textbf{Performance comparison of KG completion on FB60K-NYT10 dataset}. Results are averaged values of ten independent runs of head/tail entity predictions. The highest score is highlighted in \textbf{bold.}}
\label{tb:1}
\vskip -2ex
\end{table*}

\section{Experiment}
\subsection{Datasets and Compared Methods}
\label{sec:4.1}

\textbf{Datasets.} We use the datasets FB60K-NYT10 and UMLS-PubMed provided by \citeauthor{fu-etal-2019-collaborative}, where FB60K and UMLS are knowledge graphs and NYT10 and PubMed are corpora. FB60K-NYT10 contains more general relations (e.g., ``nationality of perso'') whereas UMLS-PubMed focuses on biomedical domain-specific relations (e.g., ``gene mapped to diseas''). We apply the pre-processed dataset \footnote{\url{https://github.com/INK-USC/CPL\#datasets}} (with training/validation/testing data size 8:1:1) to align the evaluation of our method with the baselines. Due to the imbalanced distribution and noise present in FB60K-NYT10 and UMLS-PubMed, 16 and 8 relations are selected for the performance evaluation, respectively.
We place more details of the datasets in Appendix \ref{ap:dataset}. \\
\indent\textbf{Compared Methods.} We compare our model \textsc{TagReal} with four categories of methods. For (1) traditional KG embedding-based methods, we evaluate \textbf{TransE} \cite{bordes2013translating}, \textbf{DisMult} \cite{yang2014embedding}, \textbf{ComplEx} \cite{trouillon2016complex}, \textbf{ConvE} \cite{dettmers2018convolutional}, \textbf{TuckER} \cite{balavzevic2019tucker} and \textbf{RotatE} \cite{sun2018rotate} where TuckER is a newly added model. For (2) joint text and graph embedding methods, we evaluate \textbf{RC-Net} \cite{xu2014rc}, \textbf{TransE+LINE} \cite{fu-etal-2019-collaborative} and \textbf{JointNRE} \cite{han2018neural}. For (3) reinforcement learning (RL) based path-finding methods, we evaluate \textbf{MINERVA} \cite{das2017go} and \textbf{CPL} \cite{fu-etal-2019-collaborative}. For (4) PLM-based methods, we evaluate \textbf{PKGC} \cite{lv2022pre} and our method \textbf{\textsc{TagReal}}. We keep the reported data of (2) and (3) by \citeauthor{fu-etal-2019-collaborative}\citeyear{fu-etal-2019-collaborative} while re-evaluating all models in (1) in different settings for more rigorous comparison (see Appendix \ref{ap:kge} for details). PKGC in our setting can be viewed as \textsc{TagReal} with manual prompts and without support information. 
\begin{table*}[!hbt]
\small
\centering
\setlength{\tabcolsep}{3.9pt}
\resizebox{\textwidth}{!}{
\begin{tabular}{clcccccccc}
\toprule
&
Model      & \multicolumn{2}{c}{20\%}     & \multicolumn{2}{c}{40\%}     & \multicolumn{2}{c}{70\%}     & \multicolumn{2}{c}{100\%}            \\ &      
           & Hits@5 & Hits@10             & Hits@5 & Hits@10             & Hits@5 & Hits@10             & Hits@5 & Hits@10                     \\    
\midrule
\midrule
\multirow{6}{*}{\textbf{KGE-based}} & 
TransE \cite{bordes2013translating}             & 19.70     & 30.47    & 27.72    & 41.99    & 34.62    & 49.29    & 40.83    & 53.62    \\
&
DisMult \cite{yang2014embedding}                & 19.02    & 28.35    & 28.28    & 40.48    & 32.66    & 47.01    & 39.53    & 53.82    \\
&
ComplEx \cite{trouillon2016complex}             & 11.28    & 17.17    & 24.64    & 35.15    & 25.89    & 38.19    & 34.54    & 49.30    \\
&
ConvE \cite{dettmers2018convolutional}          & 20.45    & 30.72    & 27.90    & 42.49    & 30.67    & 45.91    & 29.85    & 45.68    \\
&
TuckER \cite{balavzevic2019tucker}              & 19.94    & 30.82    & 25.79    & 41.00    & 26.48    & 42.48    & 30.22    & 45.33    \\
&
RotatE \cite{sun2018rotate}                     & 17.95    & 27.55    & 27.35    & 40.68    & 34.81    & 48.81    & 40.15    & 53.82    \\
\midrule
\multirow{3}{*}{\textbf{Text\&KGE-based}} & 
RC-Net \cite{xu2014rc}                          & 7.94     & 10.77    & 7.56     & 11.43    & 8.31     & 11.81    & 9.26     & 12.00    \\
&
TransE+Line \cite{fu-etal-2019-collaborative}   & 23.63    & 31.85    & 24.86    & 38.58    & 25.43    & 34.88    & 22.31    & 33.65    \\
&
JointNRE \cite{han2018neural}                   & 21.05    & 31.37    & 27.96    & 40.10    & 30.87    & 44.47    & -        & -      \\
\midrule
\multirow{2}{*}{\textbf{RL-based}} &
MINERVA \cite{das2017go}                        & 11.55    & 19.87    & 24.65    & 35.71    & 35.80    & 46.26    & 57.63    & 63.83    \\
&
CPL \cite{fu-etal-2019-collaborative}           & 15.32    & 24.22    & 26.96    & 38.03    & 37.23    & 47.60    & 58.10    & \textbf{65.16}    \\
\midrule
\multirow{2}{*}{\textbf{PLM-based}} & 
PKGC \cite{lv2022pre}                           & 31.08    & 43.49    & 41.34    & 52.44    & 47.39    & 55.52    & 55.05    & 59.43       \\ 
&

TagReal (our method)                       & \textbf{35.83}    & \textbf{46.45}    & \textbf{46.26}    & \textbf{55.99}    &     \textbf{53.46}    & \textbf{60.40}     & \textbf{60.68}    & 62.88       \\ 
\bottomrule
\end{tabular}}
\caption{\textbf{Performance comparison of KG completion on UMLS-PubMed dataset}. 
Results are averaged values of ten independent runs of head/tail entity predictions.  
The highest score is highlighted in
\textbf{bold.} 
}
\label{tb:2}
\end{table*}

\begin{table*}[ht]
\small
\centering
\setlength{\tabcolsep}{3.9pt}
\resizebox{0.85\textwidth}{!}{

\begin{tabular}{lccccccc}
\toprule
 Condition     & \multicolumn{3}{c}{FB60K-NYT10}     & \multicolumn{4}{c}{UMLS-PubMed}                 \\  
            \cmidrule(lr){2-4} \cmidrule(lr){5-8}
           & 20\% & 50\% & 100\%       & 20\% & 40\% & 70\% & 100\%                      \\    
\midrule
man             & (35.77, 43.82) & (41.93, 46.70) & (41.98, 52.56) & (31.08, 43.49) & (41.34, 52.44) & (47.39, 56.52) & (55.05, 59.43) \\
man+supp       & (43.23, 47.74) & (47.10, 52.02) & (48.66, 57.46) & (32.95, 44.42) & (44.37, 54.96) & (51.98, 59.09) & (59.99, 61.23)       \\
mine+supp      & (44.54, 49.53) & (47.43, 53.87) & (49.03, 58.82) & (35.56, 45.33) & (45.35, 55.44) & (53.12, 59.65) & (60.27, 61.70)\\
optim+supp    &  (45.59, 51.34) & (48.98, 55.64) & (50.85, 60.64) & (35.83, 46.45) & (46.26, 55.99) & (53.46, 60.40) & (60.68, 62.88) \\

\bottomrule
\end{tabular}}
\caption{\textbf{Ablation study on prompt and support information}. Data in brackets denotes Hits@5 (left) and Hits@10 (right). "man", "mine" and "optim" denote \textsc{TagReal} with manual prompts, mined prompt ensemble without optimization and optimized prompt ensemble, respectively. "supp" denotes application of support information.}
\vskip -1ex
\label{tb:3}
\end{table*}

\vskip -5ex
\subsection{Experimental Setup}
For FB60K-NYT10, we use LUKE \cite{yamada2020luke}, a PLM pre-trained on more Wikipedia data with RoBERTa \cite{liu2019roberta}. For UMLS-PubMed, we use SapBert \cite{liu2021self} that pre-trained on both UMLS and PubMed with BERT \cite{devlin-etal-2019-bert}. For sub-corpora mining, we use Wikipedia with 6,458,670 document examples as the general corpus and NYT10/PubMed as the reliable sources, and we mine 500 sentences at maximum (\(\theta=500\)) for each tuple. For the prompt selection, we apply MetaPAD with its default setting, and apply TruePIE with the infrequent pattern penalty, and thresholds for positive patterns and negative patterns reset to \{0.5, 0.7, 0.3\} respectively. For support information retrieval, we use BM25 to search relevant texts with \(\delta = 0.9\) and \(\phi = 100\) in the corpora NYT10/PubMed. We follow the same fine-tuning process as PKGC. We use TuckER as the KGE model to create negative triples, and we set \(M=30\) as the ratio of positive/negative triples. To compare with baselines, we test our model on training sets in the ratios of [20\%, 50\%, 100\%] for FB60K-NYT10 and [20\%, 40\%, 70\%, 100\%] for UMLS-PubMed. The evaluation metrics are described in Appendix \ref{ap:metric}.

\section{Results}
\subsection{Performance Comparison}
We show the performance comparison with the state-of-the-art methods in Tables \ref{tb:1} and \ref{tb:2}. As one can observe, \textsc{TagReal} outperforms the existing works in most cases. Given dense training data, KGE-based methods (e.g., RotatE) and RL-based methods (e.g., CPL) can still achieve relatively high performance. However, when the training data is limited, these approaches suffer, whereas PLM-based methods (PKGC and \textsc{TagReal}) are not greatly impacted. Our approach performs noticeably better in such cases than the current non-PLM-based ones. This is because the KGE models cannot be trained effectively with inadequate data, and the RL-based path-finding models cannot recognize the underlying patterns given insufficient evidential and general paths in KG. On the other hand, PLMs already possess implicit information that can be used directly, and the negative effects of insufficient data in fine-tuning would be less harsh than in training from scratch. 
\textsc{TagReal} outperforms PKGC due to its ability to automatically mine quality prompts and retrieve support information in contrast to manual annotations which are often limited. Next, we analyze the impacts of support information and prompt generation on the performance of \textsc{TagReal}.


\vskip -2em
\subsection{Model Analysis}
\label{sec:6.2}
\begin{figure}[!h]
\vskip -1ex
\advance\leftskip-0.2cm
\includegraphics[width=1.05\linewidth]{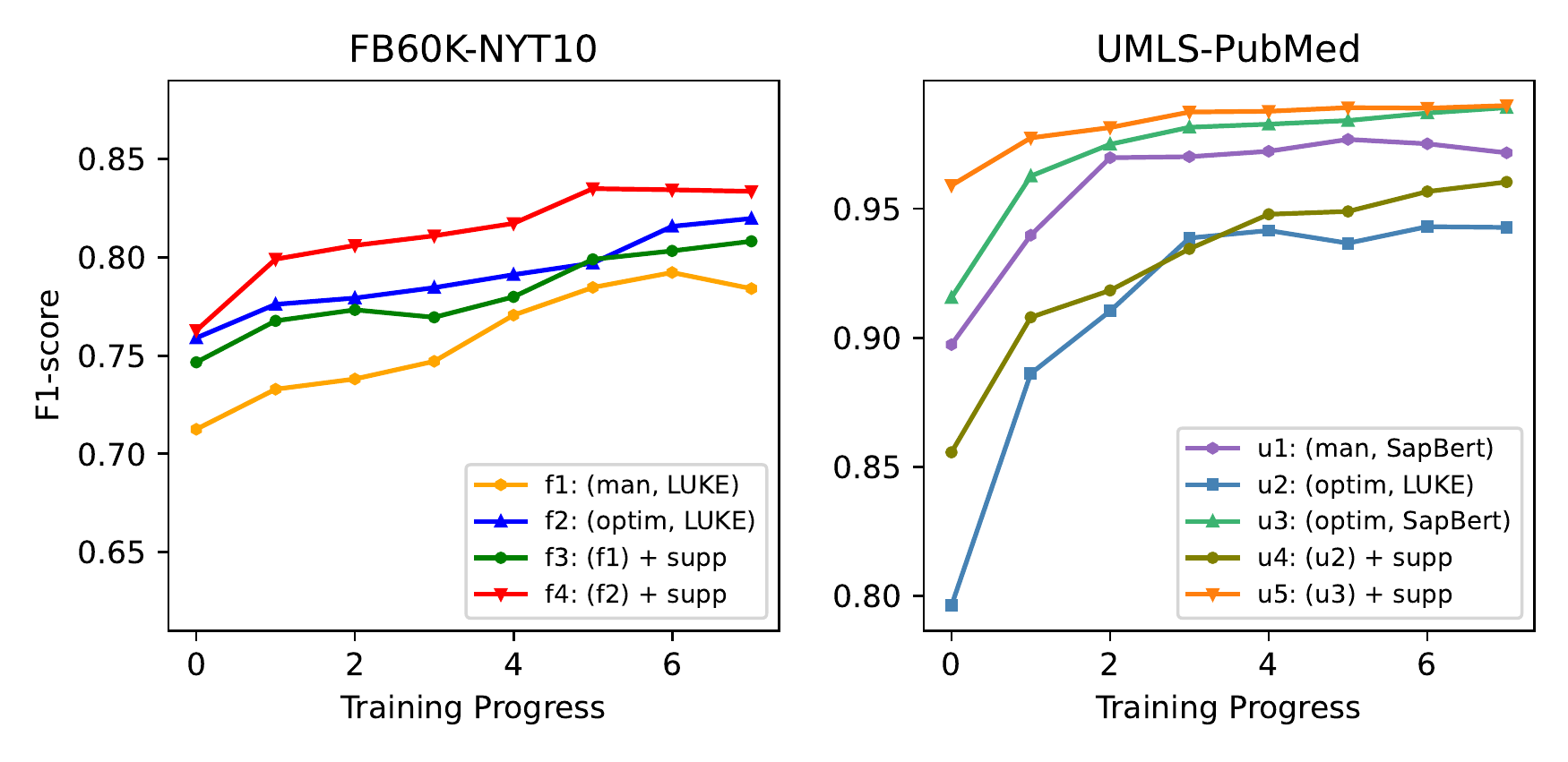}

\caption{\textbf{Performance (F1-Score) variation of triple classification w.r.t training time.} "man" or "optim" means \textsc{TagReal} with manual prompts or optimized prompt ensemble. "supp" denotes support information.
}
\vskip -2ex
\label{fig:perf_var}
\end{figure}

\begin{figure}[!h]

\centering
\includegraphics[width=0.9\linewidth]{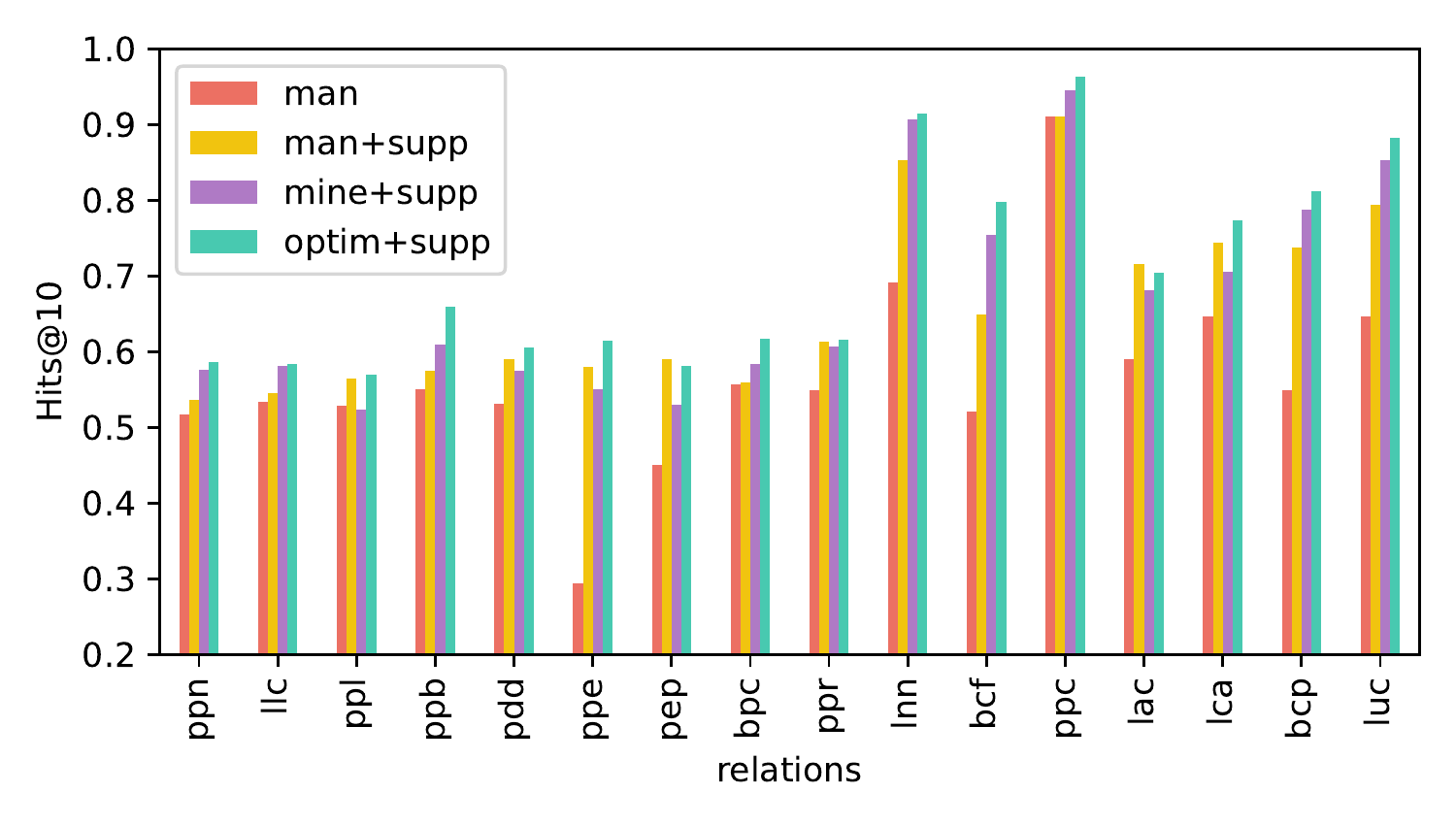}
\caption{\textbf{Relation-wise KG completion performance (Hits@10) comparison on FB60K-NYT10}. Labels on the x-axis are the abbreviations of relations.}
\label{fig:rel_wise}
\end{figure}

We conduct an ablation study to verify the effectiveness of both automatically generated prompts and retrieved support information. The results are presented in Table \ref{tb:3}, Figure \ref{fig:perf_var} and \ref{fig:rel_wise}.
\definecolor{lightgreen}{HTML}{A0E3A0}
\definecolor{lightyellow}{HTML}{FFFF00}
\definecolor{ddblue}{HTML}{71B5E8}
\begin{figure*}[!h]
  \includegraphics[width=15.91cm,height=6.4cm]{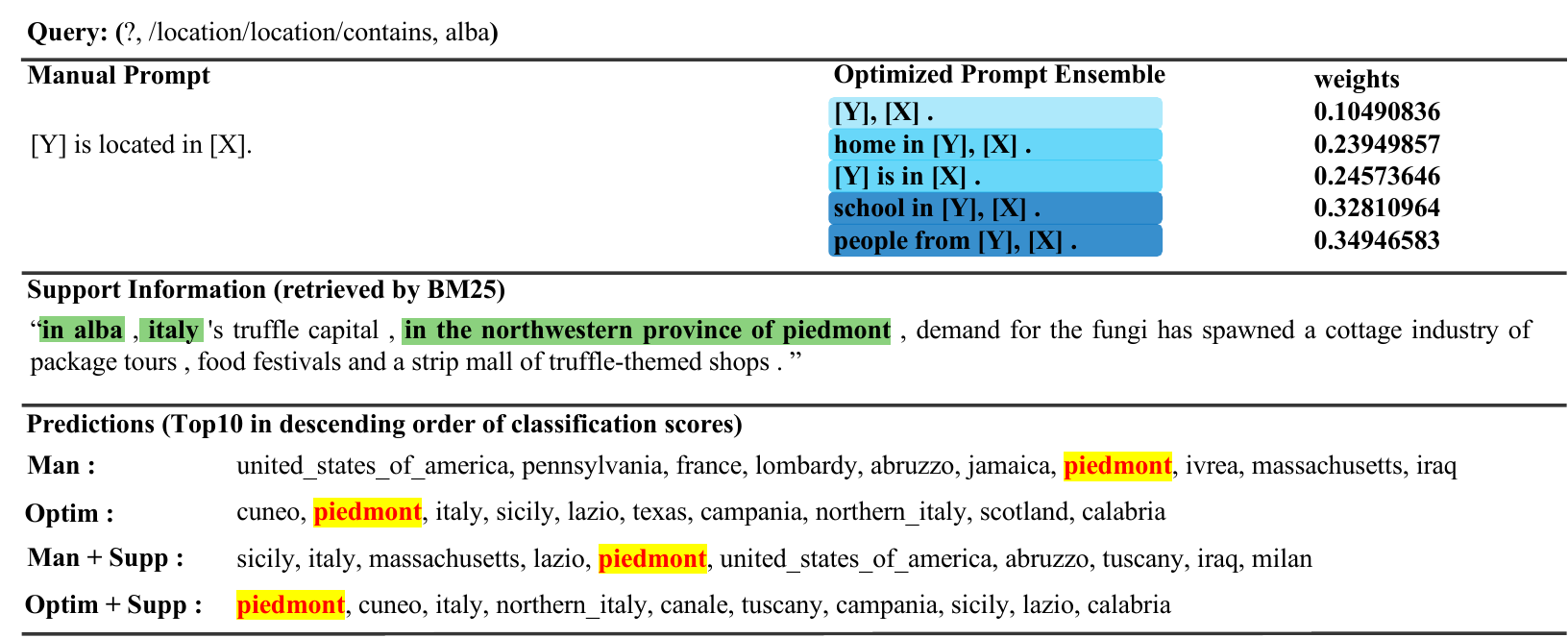}
  \caption{\textbf{Example of the link prediction with \textsc{TagReal} on FB60K-NYT10}. \textbf{Man} denotes manual prompt. \textbf{Optim} denotes optimized prompt ensemble. \textbf{Supp} denotes support information. The \colorbox{lightyellow}{\textbf{{\color{red} ground truth tail entity}}}, \colorbox{lightgreen}{\textbf{helpful information}} and \colorbox{ddblue}{\textbf{optimized prompts}} (darker for higher weights) are highlighted.}
  \label{fig:cs}
  \vskip -1ex
\end{figure*}

\textbf{Support Information.} As shown in Table \ref{tb:3}, for FB60K-NYT10, support information helps improve Hits@5 and Hits@10 in ranges of [5.2\%, 7.5\%] and [3.8\%, 5.3\%], respectively. For UMLS-PubMed, it helps improve Hits@5 and Hits10 in ranges of [1.9\%, 4.94\%] and [0.9\%, 3.6\%], respectively. Although the overlap between UMLS and PubMed is higher than that between FB60K and NYT10 \cite{fu-etal-2019-collaborative}, the textual information in PubMed could not help as much as NYT10 since that: (1) SapBert already possesses adequate implicit knowledge on both UMLS and PubMed so that a large portion of additional support texts might be useless.
The lines "u2", "u3", "u4" and "u5" in Figure \ref{fig:perf_var} show that support information helps more when using LUKE as the PLM as it contains less domain-specific knowledge. It also infers that the support information could be generalized to any application, especially when fine-tuning a PLM is difficult in low-resource scenarios \cite{arase-tsujii-2019-transfer, mahabadi2021variational}. (2) UMLS contains more queries with multiple correct answers than FB60K (see Appendix \ref{ap:dataset}), which means some queries are likely "misled" to another answer and thus not counted into the Hits@N metric.\\
\indent\textbf{Prompt Generation.} Almost all of the relations, as shown in Figure \ref{fig:rel_wise}, could be converted into better prompts by our prompt mining and optimization, albeit some of them might be marginally worse than manually created prompts due to the following fact. A few of the mined prompts, which are of lower quality than the manually created prompts, may significantly negatively affect the prediction score for the ensemble with equal weighting. Weighting based on PLM reduces such negative effects of the poor prompts for the optimized ensembles and enables them to outperform most handcrafted prompts. In addition, Table \ref{tb:3} shows the overall improvement for these three types of prompts, demonstrating that for both datasets, optimized ensembles outperform equally weighted ensembles, which in turn outperform manually created prompts. Moreover, by comparing line "f1" with line "f2", or line "u1" with line "u3" in Figure \ref{fig:perf_var}, we find a performance gap between PLM with manual prompts and with the optimized ensemble for triple classification, highlighting the effectiveness of our method.
\subsection{Case Study}
Figure \ref{fig:cs} shows an example of using \textsc{TagReal} for link prediction with a query (?\textit{, /location/location/ contains, alba}) where ``\textit{piedmont}'' is the ground truth. By comparing the prediction results in different pairs, we find that both prompt generation and support information could enhance the KG completion performance. With the handcrafted prompt, the PLM simply lists out the terms that have some connections to the subject entity ``\textit{alba}'' without being aware that we are trying to find the place it is located in. Differently, with the optimized prompt ensemble, the PLM lists entities that are highly relevant to our target, where ``\textit{cuneo}'', ``\textit{italy}'', ``\textit{northern\_italy}'' are correct real-world answers, indicating that our intention is well conveyed to the PLM. With the support information, the PLM increases the score of entities that are related to the keywords (``\textit{italy}'', ``\textit{piedmont}'') in the text. Moreover, the optimized ensemble removes ``\textit{texas}'' and ``\textit{scotland}'' from the list and leaves only Italy-related locations. More examples are placed in Appendix \ref{ap:cs}.

\section{Conclusion and Future Works}
In this study, we proposed a novel framework to exploit the implicit knowledge in PLM for open KG completion. Experimental results show that our method outperforms existing methods especially when the training data is limited. We showed that the optimized prompts with our approach outperform the handcrafted ones in PLM knowledge probing. The effectiveness of the support information retrieval to aid the prompting is also demonstrated. In the future, we may leverage QA model's power to retrieve more reliable support information. Another potential extension is to make our model more explainable by exploring path-finding tasks.


\section{Limitations}
Due to the nature of deep learning, our method is less explainable than path-finding-based KG completion methods (e.g., CPL), which provide a concrete reasoning path to the target entity. Composing the path with multiple queries might be an applicable strategy that is worthwhile to investigate in order to extend our work on the KG reasoning task. 

For the link prediction task, we adapt the ``recall and re-ranking'' strategy from PKGC \cite{lv2022pre}, which brings a trade-off between prediction efficiency and accuracy. We alleviate the issue by applying different hyper-parameters given different sizes of training data, which is discussed in detail in Appendix \ref{ap:rrk}.

As a common issue of existing KG completion models, the performance of our model also degrades when the input KG contains noisy data. The advantage of our approach in addressing this issue is that it can use both corpus-based textual information and implicit PLM knowledge to reduce noise.

\section{Ethical Statements}
In this study, we use two datasets FB60K-NYT10 and UMLS-PubMed, which include the knowledge graphs FB60K and UMLS as well as the text corpora NYT10 and PubMed. The data is all publicly available. Our task is knowledge graph completion, which is performed by finding missing facts given existing knowledge. This work is only relevant to NLP research and will not be put to improper use by ordinary people.

\section{Acknowledgements}
Research was supported in part by US DARPA KAIROS Program No. FA8750-19-2-1004 and INCAS Program No. HR001121C0165, National Science Foundation IIS-19-56151, IIS-17-41317, and IIS 17-04532, and the Molecule Maker Lab Institute: An AI Research Institutes program supported by NSF under Award No. 2019897, and the Institute for Geospatial Understanding through an Integrative Discovery Environment (I-GUIDE) by NSF under Award No. 2118329, and NSF Award SCH-2205289, SCH-2014438, IIS-2034479.



\bibliography{anthology,custom,references}

\begin{thebibliography}{49}
\expandafter\ifx\csname natexlab\endcsname\relax\def\natexlab#1{#1}\fi

\bibitem[{AlKhamissi et~al.(2022)AlKhamissi, Li, Celikyilmaz, Diab, and
  Ghazvininejad}]{alkhamissi2022review}
Badr AlKhamissi, Millicent Li, Asli Celikyilmaz, Mona Diab, and Marjan
  Ghazvininejad. 2022.
\newblock A review on language models as knowledge bases.
\newblock \emph{arXiv preprint arXiv:2204.06031}.

\bibitem[{Arase and Tsujii(2019)}]{arase-tsujii-2019-transfer}
Yuki Arase and Jun{'}ichi Tsujii. 2019.
\newblock \href {https://aclanthology.org/D19-1542} {Transfer fine-tuning: A
  {BERT} case study}.
\newblock In \emph{Proceedings of the 2019 Conference on Empirical Methods in
  Natural Language Processing and the 9th International Joint Conference on
  Natural Language Processing (EMNLP-IJCNLP)}, Hong Kong, China. Association
  for Computational Linguistics.

\bibitem[{Bala{\v{z}}evi{\'c} et~al.(2019)Bala{\v{z}}evi{\'c}, Allen, and
  Hospedales}]{balavzevic2019tucker}
Ivana Bala{\v{z}}evi{\'c}, Carl Allen, and Timothy~M Hospedales. 2019.
\newblock Tucker: Tensor factorization for knowledge graph completion.
\newblock \emph{arXiv preprint arXiv:1901.09590}.

\bibitem[{Bodenreider(2004)}]{bodenreider2004unified}
Olivier Bodenreider. 2004.
\newblock The unified medical language system (umls): integrating biomedical
  terminology.
\newblock \emph{Nucleic acids research}, 32(suppl\_1):D267--D270.

\bibitem[{Bordes et~al.(2013)Bordes, Usunier, Garcia-Duran, Weston, and
  Yakhnenko}]{bordes2013translating}
Antoine Bordes, Nicolas Usunier, Alberto Garcia-Duran, Jason Weston, and Oksana
  Yakhnenko. 2013.
\newblock Translating embeddings for modeling multi-relational data.
\newblock \emph{Advances in neural information processing systems}, 26.

\bibitem[{Brown et~al.(2020)Brown, Mann, Ryder, Subbiah, Kaplan, Dhariwal,
  Neelakantan, Shyam, Sastry, Askell et~al.}]{brown2020language}
Tom Brown, Benjamin Mann, Nick Ryder, Melanie Subbiah, Jared~D Kaplan, Prafulla
  Dhariwal, Arvind Neelakantan, Pranav Shyam, Girish Sastry, Amanda Askell,
  et~al. 2020.
\newblock Language models are few-shot learners.
\newblock \emph{Advances in neural information processing systems},
  33:1877--1901.

\bibitem[{Chang et~al.(2021)Chang, Liu, Gopalakrishnan, Hedayatnia, Zhou, and
  Hakkani-Tur}]{chang2021incorporating}
Ting-Yun Chang, Yang Liu, Karthik Gopalakrishnan, Behnam Hedayatnia, Pei Zhou,
  and Dilek Hakkani-Tur. 2021.
\newblock Incorporating commonsense knowledge graph in pretrained models for
  social commonsense tasks.
\newblock \emph{arXiv preprint arXiv:2105.05457}.

\bibitem[{Dai et~al.(2020)Dai, Wang, Xiong, and Guo}]{Dai2020ASO}
Yuanfei Dai, Shiping Wang, Neal~Naixue Xiong, and Wenzhong Guo. 2020.
\newblock A survey on knowledge graph embedding: Approaches, applications and
  benchmarks.
\newblock \emph{Electronics}, 9:750.

\bibitem[{Das et~al.(2017)Das, Dhuliawala, Zaheer, Vilnis, Durugkar,
  Krishnamurthy, Smola, and McCallum}]{das2017go}
Rajarshi Das, Shehzaad Dhuliawala, Manzil Zaheer, Luke Vilnis, Ishan Durugkar,
  Akshay Krishnamurthy, Alex Smola, and Andrew McCallum. 2017.
\newblock Go for a walk and arrive at the answer: Reasoning over paths in
  knowledge bases using reinforcement learning.
\newblock \emph{arXiv preprint arXiv:1711.05851}.

\bibitem[{Das et~al.(2018)Das, Dhuliawala, Zaheer, Vilnis, Durugkar,
  Krishnamurthy, Smola, and McCallum}]{das2018go}
Rajarshi Das, Shehzaad Dhuliawala, Manzil Zaheer, Luke Vilnis, Ishan Durugkar,
  Akshay Krishnamurthy, Alex Smola, and Andrew McCallum. 2018.
\newblock \href {https://openreview.net/forum?id=Syg-YfWCW} {Go for a walk and
  arrive at the answer: Reasoning over paths in knowledge bases using
  reinforcement learning}.
\newblock In \emph{International Conference on Learning Representations}.

\bibitem[{Dettmers et~al.(2018)Dettmers, Minervini, Stenetorp, and
  Riedel}]{dettmers2018convolutional}
Tim Dettmers, Pasquale Minervini, Pontus Stenetorp, and Sebastian Riedel. 2018.
\newblock Convolutional 2d knowledge graph embeddings.
\newblock In \emph{Proceedings of the AAAI conference on artificial
  intelligence}, volume~32.

\bibitem[{Devlin et~al.(2019)Devlin, Chang, Lee, and
  Toutanova}]{devlin-etal-2019-bert}
Jacob Devlin, Ming-Wei Chang, Kenton Lee, and Kristina Toutanova. 2019.
\newblock \href {https://doi.org/10.18653/v1/N19-1423} {{BERT}: Pre-training of
  deep bidirectional transformers for language understanding}.
\newblock In \emph{Proceedings of the 2019 Conference of the North {A}merican
  Chapter of the Association for Computational Linguistics: Human Language
  Technologies, Volume 1 (Long and Short Papers)}, pages 4171--4186,
  Minneapolis, Minnesota. Association for Computational Linguistics.

\bibitem[{Fu et~al.(2019)Fu, Chen, Qu, Jin, and
  Ren}]{fu-etal-2019-collaborative}
Cong Fu, Tong Chen, Meng Qu, Woojeong Jin, and Xiang Ren. 2019.
\newblock \href {https://doi.org/10.18653/v1/D19-1269} {Collaborative policy
  learning for open knowledge graph reasoning}.
\newblock In \emph{Proceedings of the 2019 Conference on Empirical Methods in
  Natural Language Processing and the 9th International Joint Conference on
  Natural Language Processing (EMNLP-IJCNLP)}, pages 2672--2681, Hong Kong,
  China. Association for Computational Linguistics.

\bibitem[{Hamilton et~al.(2018)Hamilton, Bajaj, Zitnik, Jurafsky, and
  Leskovec}]{hamilton2018embedding}
Will Hamilton, Payal Bajaj, Marinka Zitnik, Dan Jurafsky, and Jure Leskovec.
  2018.
\newblock Embedding logical queries on knowledge graphs.
\newblock \emph{Advances in neural information processing systems}, 31.

\bibitem[{Han et~al.(2000)Han, Pei, and Yin}]{han2000mining}
Jiawei Han, Jian Pei, and Yiwen Yin. 2000.
\newblock Mining frequent patterns without candidate generation.
\newblock \emph{ACM sigmod record}, 29(2):1--12.

\bibitem[{Han et~al.(2018)Han, Liu, and Sun}]{han2018neural}
Xu~Han, Zhiyuan Liu, and Maosong Sun. 2018.
\newblock Neural knowledge acquisition via mutual attention between knowledge
  graph and text.
\newblock In \emph{Proceedings of the AAAI Conference on Artificial
  Intelligence}, volume~32.

\bibitem[{Hao et~al.(2022)Hao, Tan, Tang, Zhang, Xing, and Hu}]{hao2022bertnet}
Shibo Hao, Bowen Tan, Kaiwen Tang, Hengzhe Zhang, Eric~P Xing, and Zhiting Hu.
  2022.
\newblock Bertnet: Harvesting knowledge graphs from pretrained language models.
\newblock \emph{arXiv preprint arXiv:2206.14268}.

\bibitem[{Hao et~al.(2017)Hao, Zhang, Liu, He, Liu, Wu, and
  Zhao}]{hao-etal-2017-end}
Yanchao Hao, Yuanzhe Zhang, Kang Liu, Shizhu He, Zhanyi Liu, Hua Wu, and Jun
  Zhao. 2017.
\newblock \href {https://doi.org/10.18653/v1/P17-1021} {An end-to-end model for
  question answering over knowledge base with cross-attention combining global
  knowledge}.
\newblock In \emph{Proceedings of the 55th Annual Meeting of the Association
  for Computational Linguistics (Volume 1: Long Papers)}, pages 221--231,
  Vancouver, Canada. Association for Computational Linguistics.

\bibitem[{Hogan et~al.(2021)Hogan, Blomqvist, Cochez, D’amato, Melo,
  Gutierrez, Kirrane, Gayo, Navigli, Neumaier, Ngomo, Polleres, Rashid, Rula,
  Schmelzeisen, Sequeda, Staab, and Zimmermann}]{10.1145/3447772}
Aidan Hogan, Eva Blomqvist, Michael Cochez, Claudia D’amato, Gerard~De Melo,
  Claudio Gutierrez, Sabrina Kirrane, Jos\'{e} Emilio~Labra Gayo, Roberto
  Navigli, Sebastian Neumaier, Axel-Cyrille~Ngonga Ngomo, Axel Polleres,
  Sabbir~M. Rashid, Anisa Rula, Lukas Schmelzeisen, Juan Sequeda, Steffen
  Staab, and Antoine Zimmermann. 2021.
\newblock \href {https://doi.org/10.1145/3447772} {Knowledge graphs}.
\newblock \emph{ACM Comput. Surv.}, 54(4).

\bibitem[{Jiang et~al.(2017)Jiang, Shang, Cassidy, Ren, Kaplan, Hanratty, and
  Han}]{jiang2017metapad}
Meng Jiang, Jingbo Shang, Taylor Cassidy, Xiang Ren, Lance~M Kaplan, Timothy~P
  Hanratty, and Jiawei Han. 2017.
\newblock Metapad: Meta pattern discovery from massive text corpora.
\newblock In \emph{Proceedings of the 23rd ACM SIGKDD International Conference
  on Knowledge Discovery and Data Mining}, pages 877--886.

\bibitem[{Jiang et~al.(2020)Jiang, Xu, Araki, and Neubig}]{jiang2020can}
Zhengbao Jiang, Frank~F Xu, Jun Araki, and Graham Neubig. 2020.
\newblock How can we know what language models know?
\newblock \emph{Transactions of the Association for Computational Linguistics},
  8:423--438.

\bibitem[{Kim et~al.(2020)Kim, Hong, Ko, and Seo}]{kim2020multi}
Bosung Kim, Taesuk Hong, Youngjoong Ko, and Jungyun Seo. 2020.
\newblock Multi-task learning for knowledge graph completion with pre-trained
  language models.
\newblock In \emph{Proceedings of the 28th International Conference on
  Computational Linguistics}, pages 1737--1743.

\bibitem[{Li et~al.(2018)Li, Jiang, Zhang, Qu, Hanratty, Gao, and
  Han}]{li2018truepie}
Qi~Li, Meng Jiang, Xikun Zhang, Meng Qu, Timothy~P Hanratty, Jing Gao, and
  Jiawei Han. 2018.
\newblock Truepie: Discovering reliable patterns in pattern-based information
  extraction.
\newblock In \emph{Proceedings of the 24th ACM SIGKDD International Conference
  on Knowledge Discovery \& Data Mining}, pages 1675--1684.

\bibitem[{Lin et~al.(2015)Lin, Liu, Sun, Liu, and Zhu}]{Lin2015LearningEA}
Yankai Lin, Zhiyuan Liu, Maosong Sun, Yang Liu, and Xuan Zhu. 2015.
\newblock Learning entity and relation embeddings for knowledge graph
  completion.
\newblock In \emph{AAAI}.

\bibitem[{Liu et~al.(2021)Liu, Shareghi, Meng, Basaldella, and
  Collier}]{liu2021self}
Fangyu Liu, Ehsan Shareghi, Zaiqiao Meng, Marco Basaldella, and Nigel Collier.
  2021.
\newblock Self-alignment pretraining for biomedical entity representations.
\newblock In \emph{Proceedings of the 2021 Conference of the North American
  Chapter of the Association for Computational Linguistics: Human Language
  Technologies}, pages 4228--4238.

\bibitem[{Liu et~al.(2019{\natexlab{a}})Liu, Ott, Goyal, Du, Joshi, Chen, Levy,
  Lewis, Zettlemoyer, and Stoyanov}]{Liu2019RoBERTaAR}
Yinhan Liu, Myle Ott, Naman Goyal, Jingfei Du, Mandar Joshi, Danqi Chen, Omer
  Levy, Mike Lewis, Luke Zettlemoyer, and Veselin Stoyanov. 2019{\natexlab{a}}.
\newblock Roberta: A robustly optimized bert pretraining approach.
\newblock \emph{ArXiv}, abs/1907.11692.

\bibitem[{Liu et~al.(2019{\natexlab{b}})Liu, Ott, Goyal, Du, Joshi, Chen, Levy,
  Lewis, Zettlemoyer, and Stoyanov}]{liu2019roberta}
Yinhan Liu, Myle Ott, Naman Goyal, Jingfei Du, Mandar Joshi, Danqi Chen, Omer
  Levy, Mike Lewis, Luke Zettlemoyer, and Veselin Stoyanov. 2019{\natexlab{b}}.
\newblock Roberta: A robustly optimized bert pretraining approach.
\newblock \emph{arXiv preprint arXiv:1907.11692}.

\bibitem[{Lv et~al.(2018)Lv, Hou, Li, and Liu}]{lv2018differentiating}
Xin Lv, Lei Hou, Juanzi Li, and Zhiyuan Liu. 2018.
\newblock Differentiating concepts and instances for knowledge graph embedding.
\newblock \emph{arXiv preprint arXiv:1811.04588}.

\bibitem[{Lv et~al.(2022)Lv, Lin, Cao, Hou, Li, Liu, Li, and Zhou}]{lv2022pre}
Xin Lv, Yankai Lin, Yixin Cao, Lei Hou, Juanzi Li, Zhiyuan Liu, Peng Li, and
  Jie Zhou. 2022.
\newblock Do pre-trained models benefit knowledge graph completion? a reliable
  evaluation and a reasonable approach.
\newblock In \emph{Findings of the Association for Computational Linguistics:
  ACL 2022}, pages 3570--3581.

\bibitem[{mahabadi et~al.(2021)mahabadi, Belinkov, and
  Henderson}]{mahabadi2021variational}
Rabeeh~Karimi mahabadi, Yonatan Belinkov, and James Henderson. 2021.
\newblock \href {https://openreview.net/forum?id=kvhzKz-_DMF} {Variational
  information bottleneck for effective low-resource fine-tuning}.
\newblock In \emph{International Conference on Learning Representations}.

\bibitem[{Petroni et~al.(2019{\natexlab{a}})Petroni, Rockt{\"a}schel, Lewis,
  Bakhtin, Wu, Miller, and Riedel}]{petroni2019language}
Fabio Petroni, Tim Rockt{\"a}schel, Patrick Lewis, Anton Bakhtin, Yuxiang Wu,
  Alexander~H Miller, and Sebastian Riedel. 2019{\natexlab{a}}.
\newblock Language models as knowledge bases?
\newblock \emph{arXiv preprint arXiv:1909.01066}.

\bibitem[{Petroni et~al.(2019{\natexlab{b}})Petroni, Rockt{\"a}schel, Riedel,
  Lewis, Bakhtin, Wu, and Miller}]{petroni-etal-2019-language}
Fabio Petroni, Tim Rockt{\"a}schel, Sebastian Riedel, Patrick Lewis, Anton
  Bakhtin, Yuxiang Wu, and Alexander Miller. 2019{\natexlab{b}}.
\newblock \href {https://doi.org/10.18653/v1/D19-1250} {Language models as
  knowledge bases?}
\newblock In \emph{Proceedings of the 2019 Conference on Empirical Methods in
  Natural Language Processing and the 9th International Joint Conference on
  Natural Language Processing (EMNLP-IJCNLP)}, pages 2463--2473, Hong Kong,
  China. Association for Computational Linguistics.

\bibitem[{Robertson et~al.(1995)Robertson, Walker, Jones, Hancock-Beaulieu,
  Gatford et~al.}]{robertson1995okapi}
Stephen~E Robertson, Steve Walker, Susan Jones, Micheline~M Hancock-Beaulieu,
  Mike Gatford, et~al. 1995.
\newblock Okapi at trec-3.
\newblock \emph{Nist Special Publication Sp}, 109:109.

\bibitem[{Shang et~al.(2018)Shang, Liu, Jiang, Ren, Voss, and
  Han}]{shang2018automated}
Jingbo Shang, Jialu Liu, Meng Jiang, Xiang Ren, Clare~R Voss, and Jiawei Han.
  2018.
\newblock Automated phrase mining from massive text corpora.
\newblock \emph{IEEE Transactions on Knowledge and Data Engineering},
  30(10):1825--1837.

\bibitem[{Shi and Weninger(2018)}]{shi2018open}
Baoxu Shi and Tim Weninger. 2018.
\newblock Open-world knowledge graph completion.
\newblock In \emph{Proceedings of the AAAI conference on artificial
  intelligence}, volume~32.

\bibitem[{Shin et~al.(2020)Shin, Razeghi, Logan~IV, Wallace, and
  Singh}]{shin-etal-2020-autoprompt}
Taylor Shin, Yasaman Razeghi, Robert~L. Logan~IV, Eric Wallace, and Sameer
  Singh. 2020.
\newblock \href {https://doi.org/10.18653/v1/2020.emnlp-main.346}
  {{A}uto{P}rompt: {E}liciting {K}nowledge from {L}anguage {M}odels with
  {A}utomatically {G}enerated {P}rompts}.
\newblock In \emph{Proceedings of the 2020 Conference on Empirical Methods in
  Natural Language Processing (EMNLP)}, pages 4222--4235, Online. Association
  for Computational Linguistics.

\bibitem[{Sun et~al.(2019)Sun, Deng, Nie, and Tang}]{sun2018rotate}
Zhiqing Sun, Zhi-Hong Deng, Jian-Yun Nie, and Jian Tang. 2019.
\newblock \href {https://openreview.net/forum?id=HkgEQnRqYQ} {Rotate: Knowledge
  graph embedding by relational rotation in complex space}.
\newblock In \emph{International Conference on Learning Representations}.

\bibitem[{Trouillon et~al.(2016{\natexlab{a}})Trouillon, Welbl, Riedel,
  Gaussier, and Bouchard}]{trouillon2016complex}
Th{\'e}o Trouillon, Johannes Welbl, Sebastian Riedel, {\'E}ric Gaussier, and
  Guillaume Bouchard. 2016{\natexlab{a}}.
\newblock Complex embeddings for simple link prediction.
\newblock In \emph{International conference on machine learning}, pages
  2071--2080. PMLR.

\bibitem[{Trouillon et~al.(2016{\natexlab{b}})Trouillon, Welbl, Riedel,
  Gaussier, and Bouchard}]{pmlr-v48-trouillon16}
Théo Trouillon, Johannes Welbl, Sebastian Riedel, Eric Gaussier, and Guillaume
  Bouchard. 2016{\natexlab{b}}.
\newblock \href {https://proceedings.mlr.press/v48/trouillon16.html} {Complex
  embeddings for simple link prediction}.
\newblock In \emph{Proceedings of The 33rd International Conference on Machine
  Learning}, volume~48 of \emph{Proceedings of Machine Learning Research},
  pages 2071--2080, New York, New York, USA. PMLR.

\bibitem[{Wang et~al.(2018)Wang, Wang, Liu, Chen, Zhang, and
  Qi}]{wang2018towards}
Meng Wang, Ruijie Wang, Jun Liu, Yihe Chen, Lei Zhang, and Guilin Qi. 2018.
\newblock Towards empty answers in sparql: approximating querying with rdf
  embedding.
\newblock In \emph{International semantic web conference}, pages 513--529.
  Springer.

\bibitem[{Wang et~al.(2017)Wang, Mao, Wang, and Guo}]{Wang2017KnowledgeGE}
Quan Wang, Zhendong Mao, Bin Wang, and Li~Guo. 2017.
\newblock Knowledge graph embedding: A survey of approaches and applications.
\newblock \emph{IEEE Transactions on Knowledge and Data Engineering},
  29:2724--2743.

\bibitem[{Xu et~al.(2014)Xu, Bai, Bian, Gao, Wang, Liu, and Liu}]{xu2014rc}
Chang Xu, Yalong Bai, Jiang Bian, Bin Gao, Gang Wang, Xiaoguang Liu, and
  Tie-Yan Liu. 2014.
\newblock Rc-net: A general framework for incorporating knowledge into word
  representations.
\newblock In \emph{Proceedings of the 23rd ACM international conference on
  conference on information and knowledge management}, pages 1219--1228.

\bibitem[{Yamada et~al.(2020)Yamada, Asai, Shindo, Takeda, and
  Matsumoto}]{yamada2020luke}
Ikuya Yamada, Akari Asai, Hiroyuki Shindo, Hideaki Takeda, and Yuji Matsumoto.
  2020.
\newblock Luke: Deep contextualized entity representations with entity-aware
  self-attention.
\newblock In \emph{EMNLP}.

\bibitem[{Yang et~al.(2014)Yang, Yih, He, Gao, and Deng}]{yang2014embedding}
Bishan Yang, Wen-tau Yih, Xiaodong He, Jianfeng Gao, and Li~Deng. 2014.
\newblock Embedding entities and relations for learning and inference in
  knowledge bases.
\newblock \emph{arXiv preprint arXiv:1412.6575}.

\bibitem[{Yao et~al.(2019)Yao, Mao, and Luo}]{yao2019kg}
Liang Yao, Chengsheng Mao, and Yuan Luo. 2019.
\newblock Kg-bert: Bert for knowledge graph completion.
\newblock \emph{arXiv preprint arXiv:1909.03193}.

\bibitem[{Yasunaga et~al.(2021)Yasunaga, Ren, Bosselut, Liang, and
  Leskovec}]{yasunaga-etal-2021-qa}
Michihiro Yasunaga, Hongyu Ren, Antoine Bosselut, Percy Liang, and Jure
  Leskovec. 2021.
\newblock \href {https://doi.org/10.18653/v1/2021.naacl-main.45} {{QA}-{GNN}:
  Reasoning with language models and knowledge graphs for question answering}.
\newblock In \emph{Proceedings of the 2021 Conference of the North American
  Chapter of the Association for Computational Linguistics: Human Language
  Technologies}, pages 535--546, Online. Association for Computational
  Linguistics.

\bibitem[{Zhang et~al.(2020)Zhang, Shen, Shang, and Han}]{zhang2020empower}
Yunyi Zhang, Jiaming Shen, Jingbo Shang, and Jiawei Han. 2020.
\newblock Empower entity set expansion via language model probing.
\newblock \emph{arXiv preprint arXiv:2004.13897}.

\bibitem[{Zhou et~al.(2020)Zhou, Dai, Chen, Zhang, Ren, Tang, He, and
  Yu}]{Zhou2020InteractiveRS}
Sijing Zhou, Xinyi Dai, Haokun Chen, Weinan Zhang, Kan Ren, Ruiming Tang,
  Xiuqiang He, and Yong Yu. 2020.
\newblock Interactive recommender system via knowledge graph-enhanced
  reinforcement learning.
\newblock \emph{Proceedings of the 43rd International ACM SIGIR Conference on
  Research and Development in Information Retrieval}.

\bibitem[{Zitnik et~al.(2018)Zitnik, Agrawal, and
  Leskovec}]{10.1093/bioinformatics/bty294}
Marinka Zitnik, Monica Agrawal, and Jure Leskovec. 2018.
\newblock \href {https://doi.org/10.1093/bioinformatics/bty294} {{Modeling
  polypharmacy side effects with graph convolutional networks}}.
\newblock \emph{Bioinformatics}, 34(13):i457--i466.

\end{thebibliography}

\clearpage
\appendix

\section{Dataset Overview}
\label{ap:dataset}
We use the datasets \textbf{FB60K-NYT10} and \textbf{UMLS-PubMed} provided by \cite{fu-etal-2019-collaborative} \footnote{\url{https://github.com/INK-USC/CPL\#datasets}}. They take the following steps to split the data: (1) split the data of each KG (FB60K or UMLS) in the ratio of 8:1:1 for training/validation/testing data. (2) For training data, they keep all triples in any relations. (3) For validation/testing data, they only keep the triples in 16/8 relations they concern (see relations in Table \ref{tb:qt_ratio_rel}). The processed data has \{train: 268280, valid: 8765, test: 8918\} for FB60K and \{train: 2030841, valid: 8756, test: 8689\} for UMLS. As for the corpora, there are 742536 and 5645558 documents in NYT10 and PubMed respectively.\\

\begin{table}[h!]
\centering

\resizebox{.45\textwidth}{!}{\begin{tabular}{lcc}
\toprule
& FB60K-NYT10     & UMLS-PubMed                 \\  
\midrule
\#query\_tail       & 57279     & 12956 \\
\#query\_head       & 23319     & 12956\\
\#triples/\#queries & 2.22      & 6.81\\

\bottomrule
\end{tabular}}
\caption{The number of queries and the ratio of triples/queries for FB60K-NYT10 and UMLS-PubMed}
\label{tb:qt_ratio}
\end{table}


\begin{table*}[ht]
\small
\centering
\setlength{\tabcolsep}{3.9pt}

\begin{tabular}{lcccccc}
\toprule 
relations & \#triples(all)     & \#queries(all)       & ratio(all)    & \#triples(test)     & \#queries(test)  & ratio(test)          \\  
\midrule 
\textbf{FB60K-NYT10} && \\
\textit{/people/person/nationality}                 & 44186     & 20215     & 2.19      & 4438      & 2282      & 1.94      \\
\textit{/location/location/contains}                & 42306     & 11971     & 3.53      & 4244      & 2373      & 1.79      \\
\textit{/people/person/place\_lived}                & 29160     & 12760     & 2.29      & 3094      & 2066      & 1.50      \\
\textit{/people/person/place\_of\_birth}            & 28108     & 16341     & 1.72      & 2882      & 2063      & 1.40      \\
\textit{/people/deceased\_person/place\_of\_death}  & 6882      & 4349      & 1.58      & 678       & 518       & 1.31      \\
\textit{/people/person/ethnicity}                   & 5956      & 2944      & 2.02      & 574       & 305       & 1.88      \\
\textit{/people/ethnicity/people}                   & 5956      & 2944      & 2.02      & 592       & 318       & 1.86      \\
\textit{/business/person/company}                   & 4334      & 2370      & 1.83      & 450       & 379       & 1.19      \\
\textit{/people/person/religion}                    & 3580      & 1688      & 2.12      & 300       & 175       & 1.71      \\
\textit{/location/neighborhood/neighborhood\_of}    & 1275      & 547       & 2.33      & 130       & 91        & 1.43      \\
\textit{/business/company/founders}                 & 904       & 709       & 1.28      & 94        & 87        & 1.08      \\
\textit{/people/person/children}                    & 821       & 711       & 1.15      & 56        & 56        & 1.00      \\
\textit{/location/administrative\_division/country} & 829       & 498       & 1.66      & 88        & 72        & 1.22      \\
\textit{/location/country/administrative\_divisions}& 829       & 498       & 1.66      & 102       & 79        & 1.29      \\
\textit{/business/company/place\_founded}           & 754       & 548       & 1.38      & 80        & 73        & 1.10      \\
\textit{/location/us\_county/county\_seat}          & 264       & 262       & 1.01      & 32        & 32        & 1.00      \\
\midrule
\textbf{UMLS-PubMed} && \\
\textit{may\_be\_treated\_by}                       & 71424     & 7703      & 9.27      & 7020      & 3118      & 2.25      \\
\textit{may\_treat}                                 & 71424     & 7703      & 9.27      & 6956      & 3091      & 2.25      \\
\textit{may\_be\_prevented\_by}                     & 10052     & 3232      & 3.11      & 1014      & 584       & 1.74      \\
\textit{may\_prevent}                               & 10052     & 3232      & 3.11      & 1034      & 586       & 1.76      \\
\textit{gene\_mapped\_to\_disease}                  & 6164      & 1732      & 3.56      & 596       & 331       & 1.80      \\
\textit{disease\_mapped\_to\_gene}                  & 6164      & 1732      & 3.56      & 652       & 357       & 1.82      \\
\textit{gene\_associated\_with\_disease}            & 536       & 289       & 1.85      & 58        & 49        & 1.18      \\
\textit{disease\_has\_associated\_gene}             & 536       & 289       & 1.85      & 48        & 41        & 1.17      \\

\bottomrule  
\end{tabular}
\caption{Number of triples (\#triples) and queries (\#queries) in relations for FB60K-NYT10 and UMLS-PubMed. Triples/queries for both head prediction and tail prediction are counted. "all" and "test" denote the whole dataset and testing data respectively.}
\vskip -1em
\label{tb:qt_ratio_rel}
\end{table*}

\textbf{Sub-training-set splitting.} To split the training data in the ratio of 20\%/50\% for FB60K-NYT10 or 20\%/40\%/70\% for UMLS-PubMed, we use the same random seeds (55, 83, 5583) as \citeauthor{fu-etal-2019-collaborative} used, and report the results in average. \\
\indent\textbf{Query-triple ratio.} Within the relations that we focus on, we calculate the ratio of the triples by the queries (including both \((h,r,?)\) and \((?,r,t)\)) to indicate the number of correct answers a query may have in average. The result is given in Table \ref{tb:qt_ratio}. For UMLS-PubMed, as the relations are symmetric in pairs, the number of queries for head and tail predictions are the same. Table \ref{tb:qt_ratio_rel} presents the counting in a more detailed setting. Both tables show that there are more multi-answer queries in UMLS-PubMed than in FB60K-NYT10, which explains why the support information may not be as helpful in the former as it is in the latter, as revealed by Table \ref{tb:3} and discussed in Section \ref{sec:6.2}.

\section{Textual Pattern Mining}
\begin{figure*}[!htp]
\centering
  \includegraphics[width=16.22cm,height=18.73cm]{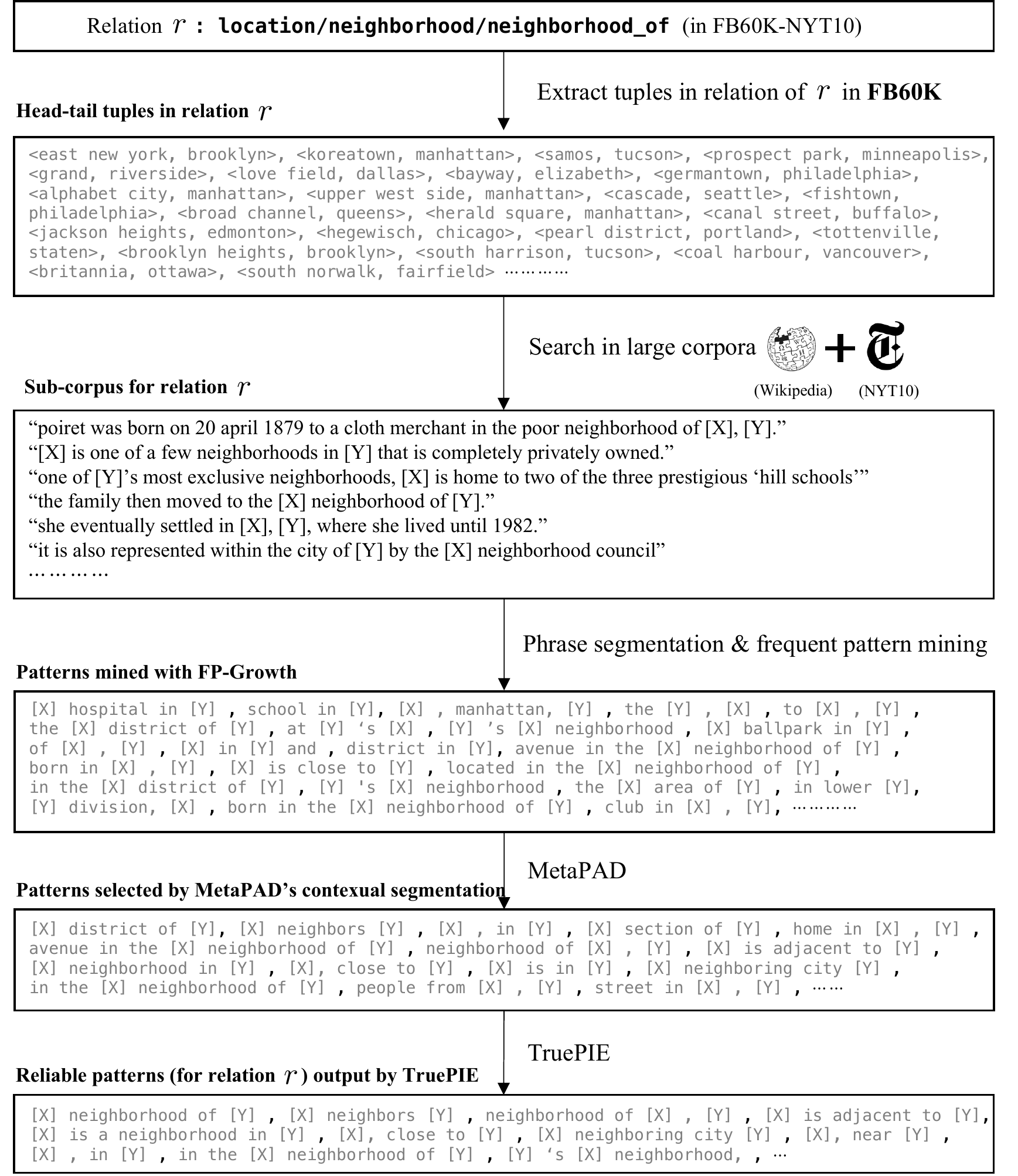}
  \caption{Example of textual pattern mining}
  \label{fig:pm_ap}
\end{figure*}
The purpose of pattern mining is to find rules that describe particular patterns in the data. Information extraction is a common goal for pattern mining and prompt mining, where the former focuses on extracting facts from massive text corpora and the latter on extracting facts from PLMs. In this section, we use another example (Figure \ref{fig:pm_ap}) to explain in detail how the textual pattern mining approaches like MetaPAD \cite{jiang2017metapad} and TruePIE \cite{li2018truepie} are implemented to mine quality prompts. In the example, given the relation \texttt{location/neighborhood/neighborhood\_of} as the input, we first extract tuples (e.g., \texttt{<east new york, brooklyn>}) in the relation from the KG (i.e., FB60K). Then, we construct a sub-corpus by searching the sentences in a large corpus (e.g., Wikipedia) and the KG-related corpus (i.e. NYT10 for FB60). After the creation of sub-corpus, we apply phrase segmentation and frequent pattern mining to mine raw prompt candidates. Since the candidate set is noisy as some prompts with low completeness (e.g., \texttt{in lower [Y]}), low informativeness (e.g., \texttt{the [Y], [X]}) and low coverage (e.g., \texttt{[X], manhattan, [Y]}) are present, we use MetaPAD to handle the prompt filtering with its quality function introducing those contextual features. After the prompts have been processed by MetaPAD, we choose one of them to serve as a seed prompt (for example, \texttt{[X] neighborhood of [Y]}) so that other prompts can be compared to it by computing their cosine similarity. As the positive seed prompt is selected manually, we can tell that there is still room for future improvement.

\section{Re-ranking Recalls from KGE Model}
\label{ap:rrk}
\textbf{Re-ranking framework.} According to the inference process we present in Figure \ref{fig:w}, we fill the placeholder (\texttt{[MASK]}) with each entity \((e_1, e_2, ..., e_n)\) in the entity set \(E\). However, as mentioned by \citeauthor{lv2022pre}\citeyear{lv2022pre}, the inference speed of PLM-based models is much slower than that of KGE models, which is a disadvantage of using PLM for KG completion. To address this issue, they use the recalls from KGE models, that is, using KGE models to run KG completion and select \(X\) top-ranked entities for each query as the entity set \(E\). Then, they shuffle the set and re-rank those entities using the PLM-based model. In our work, we adapt this re-ranking framework to accelerate the inference and evaluation as our time complexity is \(Z\) times as large as PKGC \cite{lv2018differentiating} for each case where \(Z\) is the size of prompt ensemble. We use the recalls from TuckER \cite{balavzevic2019tucker} for both datasets. \\

\textbf{Limitations.} Nonetheless, implementing the re-ranking framework has a trade-off between efficiency and Hits@N performance. When the training data is large (e.g., 100\%), the KGE model could be well trained so that the ground truth entity \(e_{gt}\) is more likely to be contained in the top \(X\) ranked ones. However, when the training data is limited (e.g., 20\%), the trained KGE model could not perform well on link prediction, as shown in Table \ref{tb:1} and \ref{tb:2}. In such a case, there is a probability that \(e_{gt}\) is not among the top \(X\) entities if we keep using the same \(X\) regardless of the size of the training data. To alleviate this side effect, we test and select different values of the hyper-parameter \(X\) for different sizes of training data, as presented in Table \ref{tb:hyper}.
\begin{table}[h!]
\centering

\resizebox{.45\textwidth}{!}{\begin{tabular}{lccccc}
\toprule
Daatset             & 20\%     & 40\%     & 50\%       & 70\%        & 100\%            \\  
\midrule
FB60K-NYT10       & 70 & -  & 40   & - & 20     \\
UMLS-PubMed       & 50 & 50 & - & 30 & 30       \\

\bottomrule
\end{tabular}}
\caption{Best \(X\) for different training sizes}
\label{tb:hyper}
\end{table}

\noindent To check how much space there is for improvement, we manually add the ground truth entity into the recalls (we should not do this for the evaluation of \textsc{TagReal} as we suppose the object entity is unknown) and test the performance of \textsc{TagReal} on UMLS-PubMed. The result is shown in Table \ref{tb:gt_recall}. By comparing this data with Table \ref{tb:3} for UMLS-PubMed, we find that changing the values of \(X\) could not perfectly address the issue. We leave the improvement as one of our major future works.


\begin{table}[ht]
\small
\centering
\setlength{\tabcolsep}{3.9pt}
\resizebox{.48\textwidth}{!}{

\begin{tabular}{lcccc}
\toprule
 Condition     & 20\% & 40\% & 70\% & 100\%                      \\    
\midrule
man        & (44.83, 60.99) & (50.81, 67.69) & (52.98, 69.21) & (60.19, 72.58)       \\
mine      & (44.98, 61.56) & (52.81, 68.66) & (56.30, 70.20) & (61.29, 74.76)\\
optim    & (45.71, 63.61) & (54.22, 69.03) & (58.18, 71.05) & (63.67, 75.55) \\

\bottomrule
\end{tabular}}
\caption{\textbf{Link prediction of \textsc{TagReal} on UMLS-PubMed with ground truth added to the KGE recalls}. Data in brackets are Hits@5 (left) and Hits@10 (right).}
\label{tb:gt_recall}
\vskip -1em
\end{table}
\section{Computing Infrastructure \& Budget}
We trained and evaluated \textsc{TagReal} on 7 NVIDIA RTX A6000 running in parallel as we support multi-GPU computing. Training \textsc{TagReal} to a good performance took about 22 and 14 hours on the entire FB60K-NYT10 dataset (with LUKE \cite{yamada2020luke}) and the entire UMLS-PubMed dataset (with SapBert \cite{liu2021self}) respectively. The training time is proportional to the size (ratio) of the training data. The evaluation took about 12 minutes for FB60K-NYT10 with LUKE when hyper-parameter \(X=20\), and 16 minutes for UMLS-PubMed with SapBert when \(X=30\). The evaluation time is proportional to \(X\), which explains why we applied the re-ranking framework (Appendix \ref{ap:rrk}) to improve the prediction efficiency. 

\section{Link Prediction with Ensemble}
\label{ap:lp}
\begin{figure}[!h]
\vskip -1em
\centering
\includegraphics[width=\linewidth]{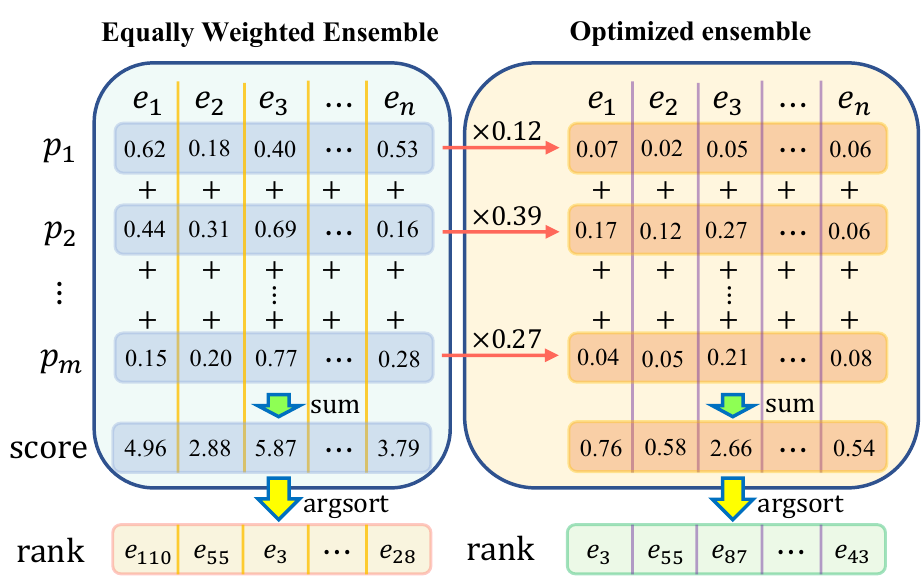}

\caption{Prediction with ensemble. \(e_1, e_2, ..., e_n \) denote the indices of entities. \(p_1, p_2, ..., p_m \) denote the prompts in the ensemble.
}
\label{fig:w}
\end{figure}
For the link prediction with equally-weighted or optimized ensembles, we apply the method shown in Figure \ref{fig:w}. Specifically, for each sentence with \texttt{[MASK]} filled with an entity \(e_i\), we calculate its classification score with the fine-tuned PLM. For each query, we get an \(m \times n\) matrix where \(m\) is the number of prompts in the ensemble, \(n\) is the number of entities in the entity set (which is \(X\) if the re-ranking framework is applied). For an ensemble that is equally weighted, we simply sum the scores of each entity obtained from the different prompts, whereas for an optimized ensemble, we multiply the weighting of the prompts by the scores before the addition. After sorting the vector in size of \(1 \times n\) in descending order, we can get the ranking of entities as the result of the link prediction.

\section{Evaluation Metrics}
\label{ap:metric}
Following previous KG completion works \cite{fu-etal-2019-collaborative, lv2022pre}, we use Hits@N and Mean Reciprocal Rank (MRR) as our evaluation metrics.
As mentioned in Section \ref{sec:3.4}, the prediction of each query \((h,r,?)\) is a 1-d vector of indices of entities in descending order regarding their scores. 
Specifically, for a query \(q_i\), We record the rank of the object entity \(t\) as \(\textsc{r}_i\), then we have:
\begin{equation}
    Hits@N = \sum_{i=1}^{Q}\frac{\textsc{r}_{i,in}}{Q} \text{ and } \\
    \textsc{r}_{i,in} =
        \begin{cases}
            0, \textsc{r}_i > N  \\
            1, \textsc{r}_i \leq N,
        \end{cases}
\end{equation}
\begin{equation}
    MRR =  \sum_{i=1}^{Q}\frac{1}{Q\textsc{r}_{i}},
\end{equation}
where \(Q\) is the number of queries in evaluation.

\section{Code Interpretation}
\begin{figure}[!h]
\vskip -1em
\centering
\includegraphics[width=0.95\linewidth]{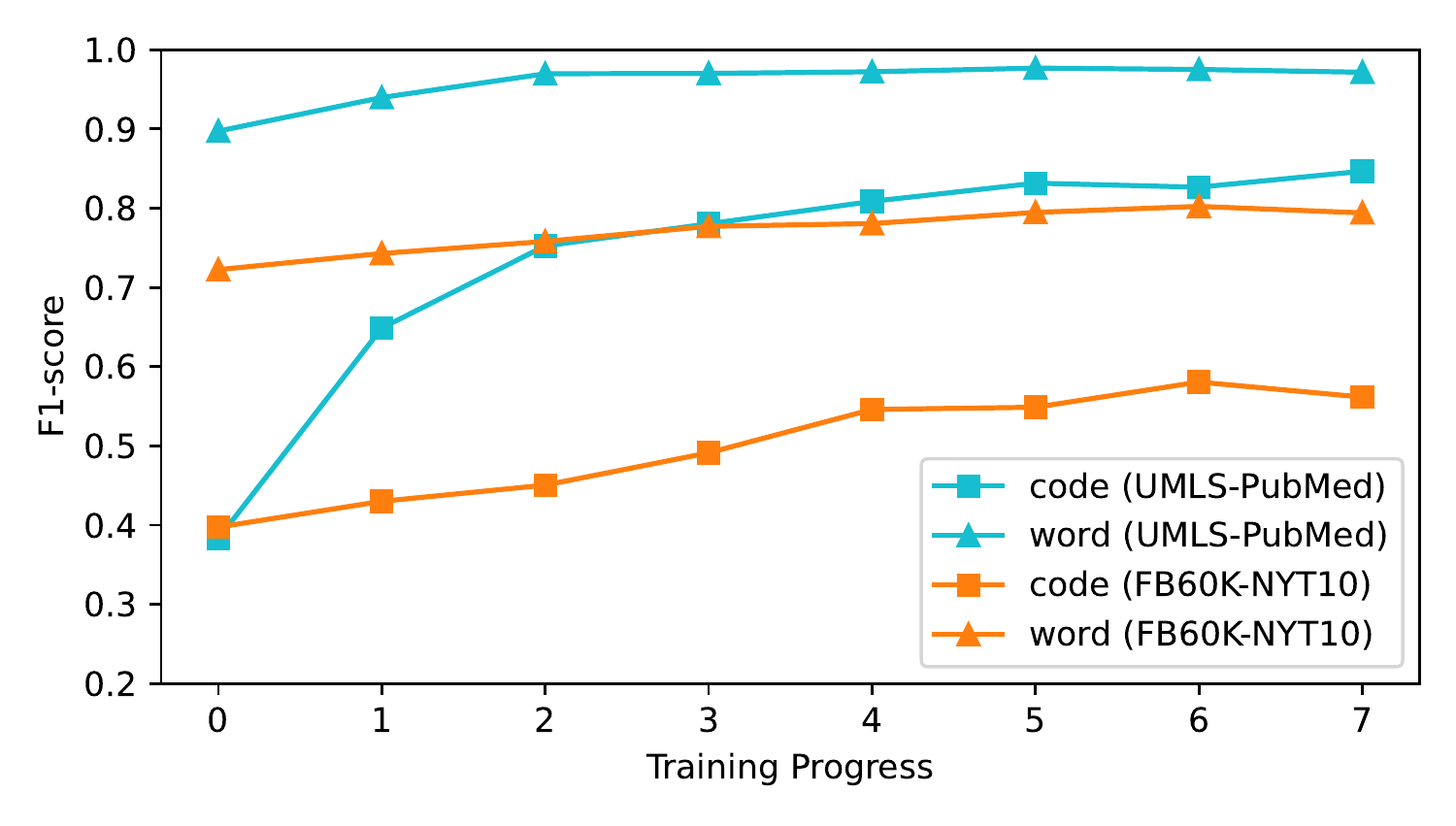}

\caption{\textbf{Performance variation of triple classification w.r.t training time.} 
"code" and "word" denote the representation of KG entities.}

\label{fig:wcp}
\end{figure}
\definecolor{lightgreen}{HTML}{A0E3A0}
\definecolor{lightyellow}{HTML}{FFFF00}
\definecolor{ddblue}{HTML}{71B5E8}
\begin{figure*}[!htp]
\centering
  \includegraphics[width=16.22cm,height=21.05cm]{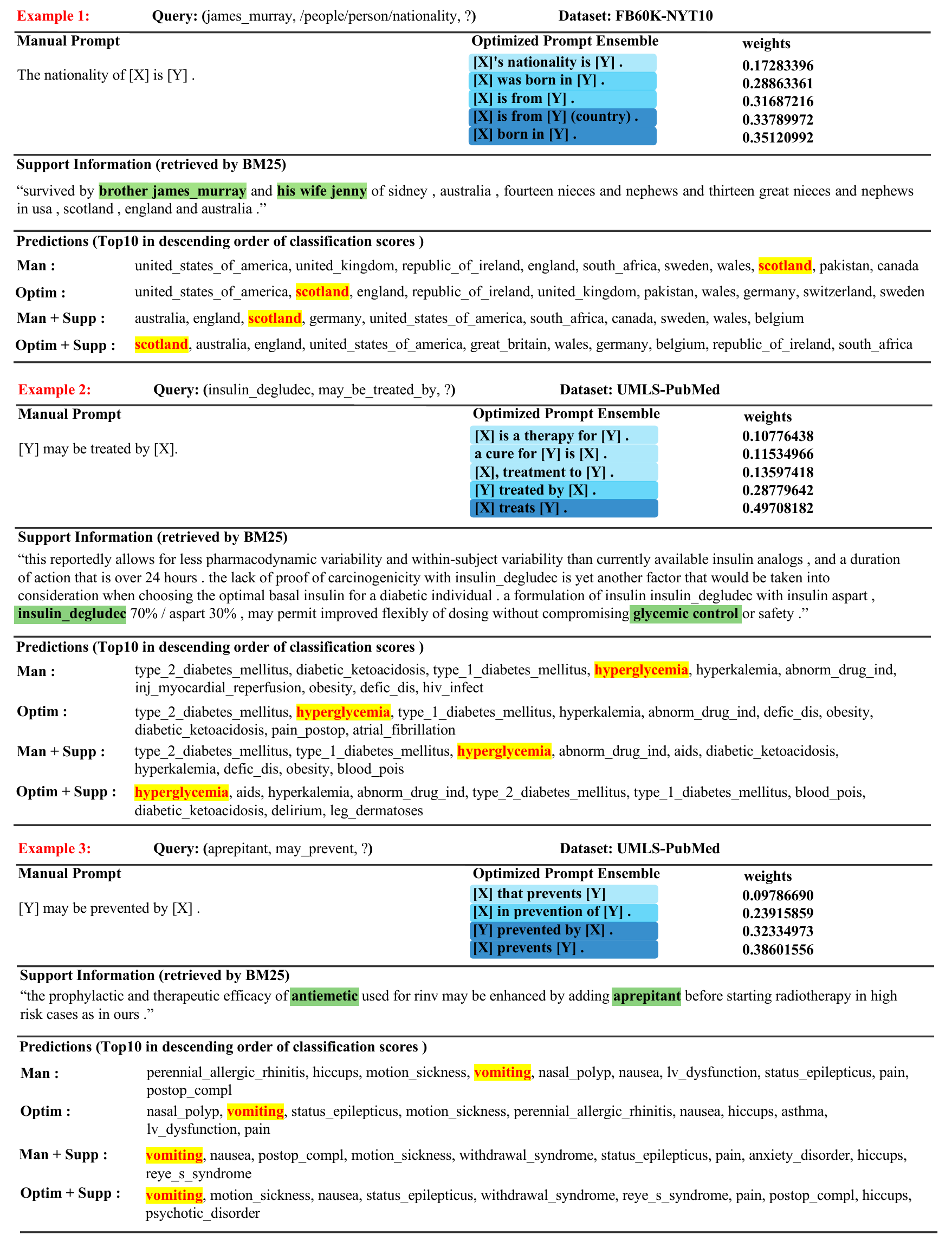}
  \caption{\textbf{Examples of the link prediction with \textsc{TagReal}}. \textbf{Man} denotes manual prompt. \textbf{Optim} denotes optimized prompt ensemble. \textbf{Supp} denotes support information. The \colorbox{lightyellow}{\textbf{{\color{red} ground truth tail entity}}}, \colorbox{lightgreen}{\textbf{helpful information}} and \colorbox{ddblue}{\textbf{optimized prompts}} (darker for higher weights) are highlighted.}
  \label{fig:cs_more}
\end{figure*}
To exploit the power of PLM, we need to map the code (entity\_id) in KG/corpus into the words (Figure \ref{fig:wcp} shows the performance difference of PLM between using word and using code). For FB60K-NYT10, we use the mapping provided by JointNRE \cite{han2018neural} \footnote{\url{https://github.com/thunlp/JointNRE}}, which covers the translation for all entities. For UMLS-PubMed, we jointly use three mappings \footnote{\url{https://evs.nci.nih.gov/ftp1/NCI_Thesaurus/}}\textsuperscript{,}\footnote{\url{https://www.ncbi.nlm.nih.gov/books/NBK9685/}}\textsuperscript{,}\footnote{\url{https://bioportal.bioontology.org/ontologies/VANDF}} which cover 97.22\% of all entities.

\section{Case Study}
\label{ap:cs}
In addition to Figure \ref{fig:cs}, we show more examples applying \textsc{TagReal} on link prediction in Figure \ref{fig:cs_more}. We can see that the predictions with optimized prompt ensemble outperform those with manual prompts in all the cases, and even outperforms predictions with manual prompts and support information in some cases. In all these examples, the support information aids the PLM knowledge probing in different ways. For the first example, we believe that the PLM captures the words ``\textit{brother james\_murray}'' and ``\textit{his wife jenny}'', and realize that we are talking about the Scottish lexicographer ``\textit{james\_murray}'' but not the American comedian with the same name, based on our survey. For the second example, the PLM probably captures ``\textit{glycemic control}'' which is highly relevant to the disease ``\textit{hyperglycemia}''. For the third example, the term ``\textit{antiemetic}'' (the drug against vomiting) is likely captured so that the answer ``\textit{vomiting}'' could be correctly predicted. Hence, it is not necessary for the support information to include the object entity itself, and including only some text relevant to it could also be helpful.

\section{Re-evaluation of Knowledge Graph Embedding Models}
We find that the performance of some KGE models was underestimated by \citeauthor{fu-etal-2019-collaborative}\citeyear{fu-etal-2019-collaborative} due to the low embedding dimension set for entity and relation. According to our re-evaluation (Table \ref{tb:kge_repro}), many of these models could perform much better with higher dimension, and we report their best performance in Table \ref{tb:1} and \ref{tb:2} based on our experiments. For the previously evaluated models, we use the same code \footnote{\url{https://github.com/thunlp/OpenKE}}\textsuperscript{,}\footnote{\url{https://github.com/DeepGraphLearning/KnowledgeGraphEmbedding}}\textsuperscript{,}\footnote{\url{https://github.com/TimDettmers/ConvE}}  as \citeauthor{fu-etal-2019-collaborative} used to ensure the fairness of the comparison. For TuckER \cite{balavzevic2019tucker}, we use the code provided by the author \footnote{\url{https://github.com/ibalazevic/TuckER}}. Same as \citeauthor{fu-etal-2019-collaborative}, to make the comparison more rigorous, we do not apply the filtered setting \cite{bordes2013translating, sun2018rotate} of the Hits@N evaluation to all the models including \textsc{TagReal}.
\label{ap:kge}

\section{Code for Paper}
The source code of \textsc{TagReal} is attached as supplemental material to the submission of this paper.
\begin{table*}[!h]
\small
\centering
\resizebox{1.0\textwidth}{!}{

\begin{tabular}{lllll}
\toprule
\textbf{FB60K-NYT10}     & \multicolumn{2}{c}{\citeauthor{fu-etal-2019-collaborative}'s setting}     & \multicolumn{2}{c}{Our setting}            \\       
& (edim, rdim, filter) & Ratio: (Hits@5, Hits@10, MRR)        & (edim, rdim, filter) & Ratio: (Hits@5, Hits@10, MRR)               \\    
\midrule
\midrule
TransE \cite{bordes2013translating}             & (100, 100, n/a)   & 20\%: (15.12, 18.83, 12.57)   & \textbf{(600, 600, n/a)}   & 20\%: (29.13, 32.67, 15.80)        \\
&& 50\%: (19.38, 23.20, 13.36) && 50\%: (41.54, 45.74. 25.82)         \\
&& 100\%: (38.53, 43.38, 29.90) && 100\%: (42.53, 46.77, 29.86)          \\
\midrule                                        
DisMult \cite{yang2014embedding}                & (100, 100, n/a)    & 20\%: (1.42, 2.55, 1.05)    & \textbf{(600, 600, n/a)}    & 20\%: (3.44, 4.31, 2.64)        \\
&& 50\%: (15.23, 19.05, 12.36)  &&  50\%: (15.98, 18.85, 13.14)         \\
&& 100\%: (32.11, 35.88, 24.95) && 100\%: (37.94, 41.62, 30.56)         \\
\midrule
ComplEx \cite{trouillon2016complex}             & (100, 100, n/a)    & 20\%: (4.22, 5.97, 3.44)    & \textbf{(600, 600, n/a)}    & 20\%: (4.32, 5.48, 3.16)        \\
&& 50\%: (19.10, 23.08, 12.99)  &&  50\%: (15.00, 17.73, 12.21)         \\
&& 100\%: (32.91, 34.62, 24.67) && 100\%: (35.42, 38.85, 28.59)         \\
\midrule
ConvE \cite{dettmers2018convolutional}          & (200, 200, n/a)    & 20\%: (20.60, 26.90, 11.96)    & (100, 100, n/a)    & 20\%: (22.91, 26.29, 19.48)        \\
&& 50\%: (24.39, 30.59, 18.51)  &&  50\%: (26.52, 29.84, 22.67)         \\
&& 100\%: (33.02, 39.78, 24.45) && 100\%: (31.71, 35.66, 25.58)         \\
\cmidrule{4-5}
&&& \textbf{(600, 600, n/a)} & 20\%: (29.49, 33.30, 24.31) \\
&&&& 50\%: (40.10, 44.03, 32.97) \\
&&&& 100\%: (50.18, 54.06, 40.39) \\
\midrule
TuckER \cite{balavzevic2019tucker}              & - & - & (100, 100, n/a)    & 20\%: (20.04, 23.02, 16.27)        \\
&&&&  50\%: (24.04, 27.88, 20.21)         \\
&&&& 100\%: (34.54, 38.77, 28.19)         \\
\cmidrule{4-5}
&&& \textbf{(600, 600, n/a)} & 20\%: (29.50, 32.48, 24.44) \\
&&&& 50\%: (41.73, 45.58, 33.84) \\
&&&& 100\%: (51.09, 54.80, 40.47) \\
\midrule
RotatE \cite{sun2018rotate}                     & (200, 100, ?)    & 20\%: (9.25, 11.83, 8.04)    & (100, 50, n/a)    & 20\%: (1.34, 2.13, 1.08)        \\
&& 50\%: (25.96, 31.63, 23.34)  &&  50\%: (2.54, 4.03, 1.91)         \\
&& 100\%: (58.32, 60.66, 51.85) && 100\%: (5.42, 7.87, 2.09)         \\
\cmidrule{4-5}
&&& (200, 100, n/a) & 20\%: (7.47, 9.14, 5.81) \\
&&&& 50\%: (21.68, 25.45, 17.35) \\
&&&& 100\%: (47.96, 52.02, 39.17) \\
\cmidrule{4-5}
&&& \textbf{(600, 300, n/a)} & 20\%: (15.91, 18.32, 12.65) \\
&&&& 50\%: (35.48, 39.42, 28.92) \\
&&&& 100\%: (51.73, 55.27, 42.64) \\
\midrule
\cr
\midrule
\textbf{UMLS-PubMed}

& \multicolumn{2}{c}{\citeauthor{fu-etal-2019-collaborative}'s setting}     & \multicolumn{2}{c}{Our setting}            \\       
& (edim, rdim, filter) & Ratio: (Hits@5, Hits@10)        & (edim, rdim, filter) & Ratio: (Hits@5, Hits@10)               \\    
\midrule
\midrule
TransE \cite{bordes2013translating}             & (100, 100, n/a)   & 20\%: (7.12, 11.17)   & \textbf{(600, 600, n/a)}   & 20\%: (19.70, 30.47)        \\
&& 40\%: (26.86, 38.08) && 40\%: (27.72, 41.99)         \\
&& 70\%: (31.32, 43.58) && 70\%: (34.62, 49.29)         \\
&& 100\%: (32.28, 45.52) && 100\%: (40.83, 53.62)          \\
\midrule  
DisMult \cite{yang2014embedding}             & (100, 100, n/a)   & 20\%: (14.66, 21.16)   & \textbf{(600, 600, n/a)}   & 20\%: (19.02, 28.35)        \\
&& 40\%: (26.90, 38.35) && 40\%: (28.28, 40.48)         \\
&& 70\%: (31.65, 44.98) && 70\%: (32.66, 47.01)         \\
&& 100\%: (32.80, 47.50) && 100\%: (39.53, 53.82)          \\
\midrule
ComplEx \cite{trouillon2016complex}             & (100, 100, n/a)   & 20\%: (18.18, 19.58)   & \textbf{(600, 600, n/a)}   & 20\%: (11.28, 17.17)        \\
&& 40\%: (23.77, 34.15) && 40\%: (24.64, 35.15)         \\
&& 70\%: (30.04, 43.60) && 70\%: (25.89, 38.19)         \\
&& 100\%: (31.84, 46.57) && 100\%: (34.54, 49.30)          \\
\midrule
ConvE \cite{dettmers2018convolutional}          & (200, 200, n/a)   & 20\%: (20.51, 30.11)   & \textbf{(200, 200, n/a)}   & 20\%: (20.45, 30.72)        \\
&& 40\%: (28.01, 42.04) && 40\%: (27.90, 42.49)         \\
&& 70\%: (31.01, 45.81) && 70\%: (30.67, 45.91)         \\
&& 100\%: (30.35, 45.35) && 100\%: (29.85, 45.68)          \\
\cmidrule{4-5}
&&& (600, 600, n/a) & 20\%: (20.26, 30.29) \\
&&&& 40\%: (26.85, 41.57) \\
&&&& 70\%: (26.97, 42.44) \\
&&&& 100\%: (25.43, 41.58) \\
\midrule
TuckER \cite{balavzevic2019tucker}              
& - & - & (100, 100, n/a)    & 20\%: (5.13, 8.06)        \\
&&&&  40\%: (20.48, 31.20)         \\
&&&&  70\%: (29.66, 42.89)          \\
&&&& 100\%: (31.56, 44.72)         \\
\cmidrule{4-5}
&&& \textbf{(256, 256, n/a)} & 20\%: (19.94, 30.82) \\
&&&& 40\%: (25.79, 41.00) \\
&&&& 70\%: (26.48, 42.48)               \\
&&&& 100\%: (30.22, 45.33) \\
\cmidrule{4-5}
&&& (600, 600, n/a) & 20\%: (18.84, 27.94) \\
&&&& 40\%: (24.57, 37.79) \\
&&&& 70\%: (25.50, 41.32)  \\
&&&& 100\%: (24.41, 40.56) \\

\midrule
RotatE \cite{sun2018rotate}                     
& (200, 100, n/a)    
& 20\%: (4.03, 6.50)    
& \textbf{(600, 300, n/a)}    
& 20\%: (17.95, 27.55)        \\
&& 40\%: (8.65, 13.21)  
&&  40\%: (27.35, 40.68)         \\
&& 70\%: (14.90, 21.67) 
&& 70\%: (34.81, 48.81) \\
&& 100\%: (20.75, 27.82) 
&& 100\%: (40.15, 53.82) \\

\bottomrule
\end{tabular}
}
\caption{\textbf{Performance of knowledge graph embedding models on FB60K-NYT10 and UMLS-PubMed}. "edim" and "rdim" denotes the embedding size of entity and relation respectively. "filter" denotes the application of the filtered setting. Best setting for each model is highlighted in \textbf{bold}.}
\label{tb:kge_repro}
\end{table*}

\end{document}